\makeatletter
\def\thanks#1{\protected@xdef\@thanks{\@thanks
\protect\footnotetext{#1}}}
\makeatother

\documentclass[runningheads]{llncs}

\usepackage{eccv}

\usepackage{eccvabbrv}

\usepackage{graphicx}
\usepackage{booktabs}
\usepackage{multirow}
\usepackage[accsupp]{axessibility}  % Improves PDF readability for those with disabilities.

\usepackage{hyperref}

\usepackage{orcidlink}

\newcommand{\rxl}[1]{\textcolor{black}{#1}}

\begin{document}

% ---------------------------------------------------------------
% TODO REVIEW: Replace with your title
\title{Improving Neural Surface Reconstruction with Feature Priors from Multi-View Images}
\titlerunning{Improving Neural Surface Reconstruction with Feature Priors }

% TODO REVIEW: If the paper title is too long for the running head, you can set
% an abbreviated paper title here. If not, comment out.
% \titlerunning{Abbreviated paper title}

% TODO FINAL: Replace with your author list. 
% Include the authors' OCRID for the camera-ready version, if at all possible.

\author{Xinlin Ren\inst{1*}\thanks{*: Equal Contributions.}\orcidlink{0000-0002-8175-7392} \and
Chenjie Cao\inst{1,2*}\orcidlink{0000-0003-3916-2843} \and
Yanwei Fu\inst{1^\dagger}\thanks{$^\dagger$: Corresponding Author}\orcidlink{0000-0002-6595-6893} \and
Xiangyang Xue\inst{1}\orcidlink{0000-0002-4897-9209}}

% TODO FINAL: Replace with an abbreviated list of authors.
\authorrunning{X. Ren et al.}
% First names are abbreviated in the running head.
% If there are more than two authors, 'et al.' is used.

% TODO FINAL: Replace with your institution list.

\institute{Fudan University\\ 
 \and Alibaba Group, DAMO Academy \\
 \email{ \{xlren20,cjcao20,yanweifu,xyxue\}@fudan.edu.cn }}

% \institute{Princeton University, Princeton NJ 08544, USA \and
% Springer Heidelberg, Tiergartenstr.~17, 69121 Heidelberg, Germany
% \email{lncs@springer.com}\\
% \url{http://www.springer.com/gp/computer-science/lncs} \and
% ABC Institute, Rupert-Karls-University Heidelberg, Heidelberg, Germany\\
% \email{\{abc,lncs\}@uni-heidelberg.de}}

\maketitle

\begin{abstract}
  Recent advancements in Neural Surface Reconstruction (NSR) have significantly improved multi-view reconstruction when coupled with volume rendering. However, relying solely on photometric consistency in image space falls short of addressing complexities posed by real-world data, including occlusions and non-Lambertian surfaces.
To tackle these challenges, we propose an investigation into feature-level consistent loss, aiming to harness valuable feature priors from diverse pretext visual tasks and overcome current limitations. It is crucial to note the existing gap in determining the most effective pretext visual task for enhancing NSR.
In this study, we comprehensively explore multi-view feature priors from seven pretext visual tasks, comprising thirteen methods. Our main goal is to strengthen NSR training by considering a wide range of possibilities. Additionally, we examine the impact of varying feature resolutions and evaluate both pixel-wise and patch-wise consistent losses, providing insights into effective strategies for improving NSR performance. By incorporating pre-trained representations from MVSFormer~\cite{cao2022mvsformer} and QuadTree~\cite{tang2022quadtree}, our approach can generate variations of MVS-NeuS and Match-NeuS, respectively.
Our results, analyzed on DTU and EPFL datasets, reveal that feature priors from image matching and multi-view stereo outperform other pretext tasks. Moreover, we discover that extending patch-wise photometric consistency to the feature level surpasses the performance of pixel-wise approaches. These findings underscore the effectiveness of these techniques in enhancing NSR outcomes. Codes and models are available at \url{https://github.com/maybeLx/MVS_NeuS}.
  % \and Multi-View Stereo \and Image Matching}
\end{abstract}

\section{Introduction}
\label{sec:intro}

% \begin{figure}
% \begin{centering}
% \begin{tabular}{cc}
% \begin{tabular}{c}
% \includegraphics[width=0.6\columnwidth]{pics/teaser.pdf}\tabularnewline
% \end{tabular} & %

% \begin{tabular}{c}
% \includegraphics[width=0.35\columnwidth]{pics/feature.pdf}\tabularnewline
% \end{tabular}\tabularnewline
% {\small (a)  }  &  {\small (b) }\tabularnewline
% \end{tabular}\caption{\small (a) Pixel-wise similarity for RGB images: MVS features~\cite{cao2022mvsformer} and matching features~\cite{tang2022quadtree} on non-Lambertian surfaces. (b) Multi-view consistency by feature priors: Given a ray direction, the geometry network can extract surface points, which are used to project and align source features to the reference view for optimizing patch-wise photometric consistency loss.}
% \label{fig:teaser}
% \par\end{centering}
% \end{figure}

\begin{figure}[htb]
    \centering
    \begin{subfigure}{0.6\columnwidth}
        \includegraphics[width=1\columnwidth]{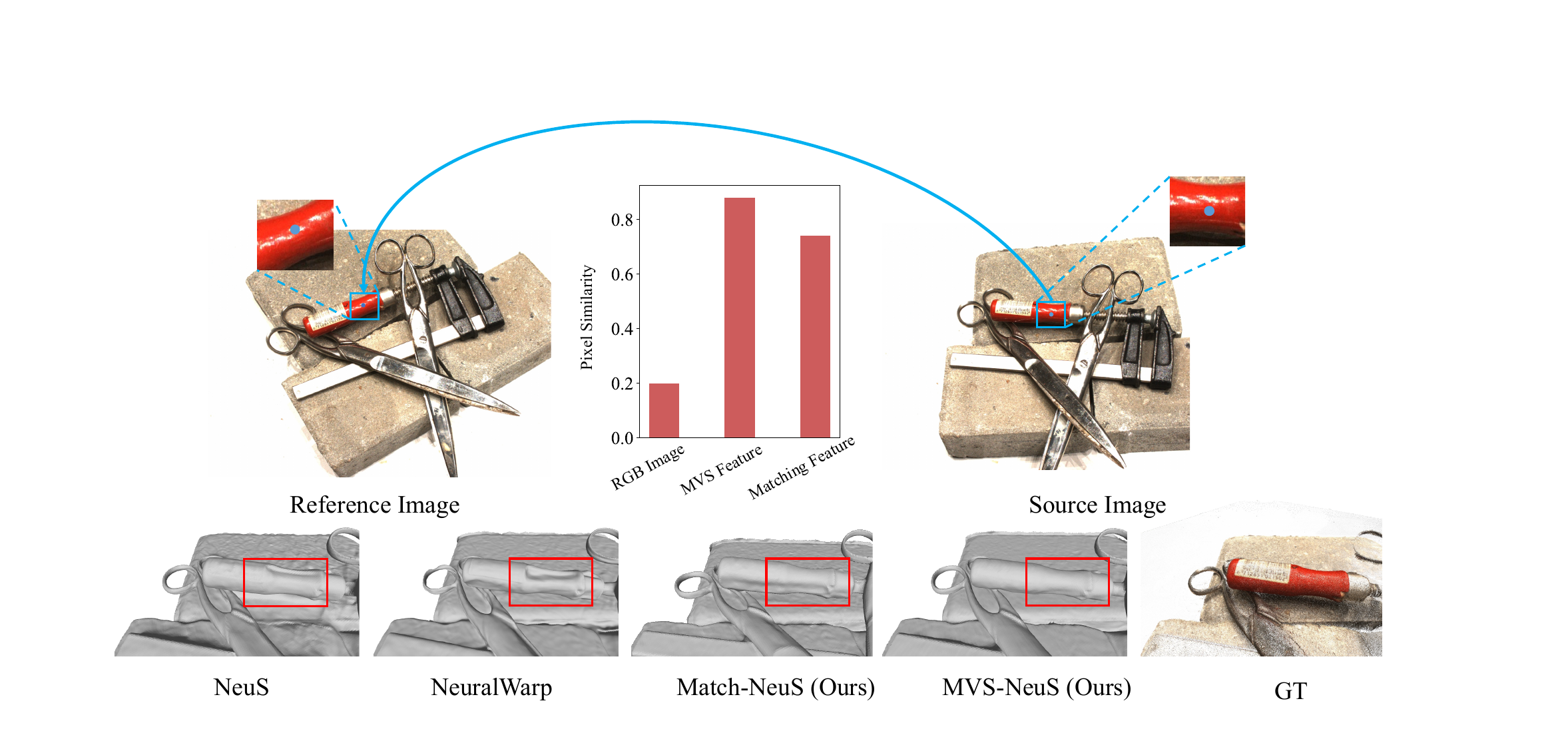}\tabularnewline
        \caption{\small}\label{fig:teaser}
    \end{subfigure}
    \begin{subfigure}{0.35\columnwidth}
        \includegraphics[width=1\columnwidth]{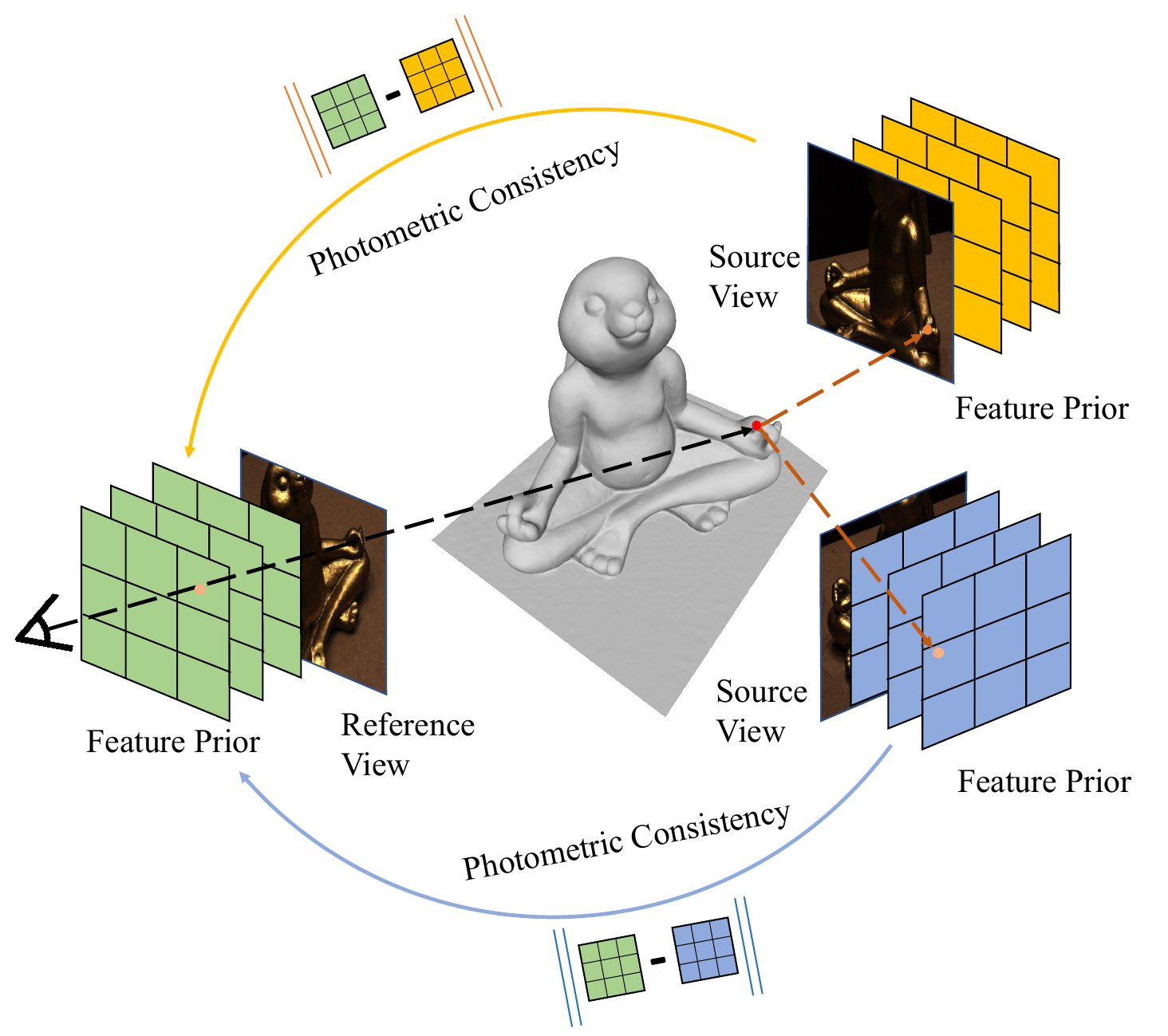}\tabularnewline
        \caption{\small}\label{fig:mult-view-consistency}
    \end{subfigure}
    \vspace{-0.1in}
    \caption{\small (a) Pixel-wise similarity for RGB images: MVS features~\cite{cao2022mvsformer} and matching features~\cite{tang2022quadtree} on non-Lambertian surfaces. (b) Multi-view consistency of feature priors: Given a ray direction, the geometry network can extract surface points, which are used to project and align source features to the reference view for optimizing patch-wise photometric consistency loss.}
    \label{fig:teaser_whole}
    \vspace{-0.2in}
\end{figure}

% \begin{figure}
%   \vspace{-0.12in}
%   \centering
%   \includegraphics[width=0.85\columnwidth]{pics/teaser.pdf}
%   % \vspace{-0.2in}
%     \vspace{-0.1in}
%   \caption{The visualization of the pixel-wise similarity for RGB images, MVS features~\cite{cao2022mvsformer} and matching features~\cite{tang2022quadtree} on non-Lambertian surfaces. }
%   \label{fig:teaser}
%   \vspace{-0.2in}
% \end{figure}

% \begin{figure}
% \vspace{-0.15in}
% \centering
% \includegraphics[width=0.46\columnwidth]{pics/feature.pdf}
%     % \vspace{0.2in}
%   \caption{\small Illustration of multi-view consistency based on feature priors. Given a ray direction, the geometry network can extract surface points, which are used to project and align source features to the reference view for optimizing patch-wise photometric consistency loss.  } 
%   \label{fig:mult-view-consistency}
%   \vspace{-0.25in}
% \end{figure}

Reconstructing surface from multi-view images is a crucial task in 3D computer vision. Traditionally, this task is addressed using a pipeline that includes depth prediction, point cloud reconstruction, and maximizing point consistency for surface mesh~\cite{labatut2007efficient, barnes2009patchmatch, kazhdan2013screened, schonberger2016pixelwise}. Unfortunately, these traditional methods often result in visual errors due to the complexity of these procedures.
In contrast, recent approaches opt to represent the surface through volume rendering, inspired by the Neural Radiance Fields (NeRF)~\cite{mildenhall2021nerf}, thereby avoiding the accumulation of errors seen in the traditional pipeline. Notably, Neural Surface Reconstruction (NSR) has made significant strides in multi-view reconstruction, leveraging techniques such as Signed Distance Functions (SDF)~\cite{yariv2021volume, wang2021neus} and occupancy~\cite{niemeyer2020differentiable} combined with volume rendering.
However, despite these advancements, these sophisticated methods still grapple with fundamental challenges in surface reconstruction, including maintaining geometric consistency across different views. This challenge becomes particularly pronounced when dealing with non-Lambertian surfaces and occlusions~\cite{darmon2022improving, fu2022geo}, as illustrated in Fig.~\ref{fig:teaser}.

To tackle these challenges, earlier studies~\cite{darmon2022improving, fu2022geo} have delved into optimizing patch-wise warping losses in image space, employing metrics such as Structural Similarity (SSIM)~\cite{wang2004image} or Normalized Cross-Correlation (NCC)~\cite{galliani2015massively}.
In an effort to enhance the robustness of geometric consistency in real-world scenarios, Chen \etal~\cite{chen2023recovering} utilized features extracted from a learning-based Multi-View Stereo (MVS) model~\cite{zhang2020visibility} to compute pixel-wise feature similarity. Despite promising results with MVS features, it remains unclear \textit{whether the neural surface reconstruction can benefit from other pretext tasks and what key factors contribute to improved reconstruction}. 
% Additionally, it is worthwhile to investigate whether reconstruction losses based on features can also benefit from patch-wise photometric consistency.
Furthermore, it is worth investigating whether these features can also benefit from patch-wise photometric consistency.

This study aims to address these questions by conducting a comprehensive investigation into the impact of diverse pre-trained feature priors obtained from various pretext visual tasks for NSR learning. We thoroughly consider seven distinctly different pretext visual tasks, encompassing 2D supervised~\cite{chu2021twins, liu2022convnet, xie2021segformer}, 2D self-supervised~\cite{he2022masked, gao2022convmae, weinzaepfel2022croco}, and 3D supervised~\cite{ranftl2020towards, kim2022global, xu2023iterative, lipson2021raft, tang2022quadtree, gu2020cascade, cao2022mvsformer} approaches.
The pre-trained models from these tasks can serve as feature extractors for our NSR task. Crucially, we not only compare various architectures used in these models, ranging from Convolution Neural Networks (CNNs)~\cite{liu2022convnet, xu2023iterative, lipson2021raft, gu2020cascade} to Vision Transformers (ViTs)~\cite{xie2021segformer, he2022masked, weinzaepfel2022croco, cao2022mvsformer}, but also explore feature maps from different layers within the same model to identify common factors relevant to NSR.
Furthermore, we propose for extending patch-wise photometric consistency to the feature level, resulting in substantial improvements.

% \begin{wrapfigure}{h}{0.5\columnwidth}
%   % \vspace{-0.12in}
%   \centering
%   \includegraphics[width=0.475\columnwidth]{pics/feature.pdf}
%     \vspace{-0.1in}
%   \caption{Illustration of multi-view consistency based on feature priors. Given a ray direction, the geometry network can extract surface points, which are used to project and align source features to the reference view for optimizing patch-wise photometric consistency loss.} 
%   \label{fig:mult-view-consistency}
%   \vspace{-0.15in}
% \end{wrapfigure}

Formally, our approach enhances NSR by integrating various feature priors derived from multi-view images. Building upon the robust foundation of NeuS~\cite{wang2021neus}, which acts as the baseline method in this paper, we further improve NSR by introducing feature-level consistent loss from diverse pre-trained models, as depicted in Fig.~\ref{fig:mult-view-consistency}. Specifically, we compare 13 pre-trained models across 7 distinct pretext tasks, as listed in Tab.~\ref{tab:pretext_tasks}. To the best of our knowledge, this evaluation represents the most comprehensive assessment of pretext tasks on NSR to date, encompassing masked image modeling \cite{gao2022convmae, he2022masked, weinzaepfel2022croco}, image classification \cite{chu2021twins, liu2022convnet}, semantic segmentation \cite{xie2021segformer}, monocular depth estimation \cite{ranftl2020towards, kim2022global}, stereo matching \cite{xu2023iterative, lipson2021raft}, image matching \cite{tang2022quadtree}, and MVS \cite{gu2020cascade, cao2022mvsformer}. The DTU and EPFL datasets serve as the benchmark for evaluation~\cite{jensen2014large}.

% \begin{figure}
% \vspace{-0.15in}
% \centering
% \includegraphics[width=0.46\columnwidth]{pics/feature.pdf}
%     % \vspace{0.2in}
%   \caption{\small Illustration of multi-view consistency based on feature priors. Given a ray direction, the geometry network can extract surface points, which are used to project and align source features to the reference view for optimizing patch-wise photometric consistency loss.  } 
%   \label{fig:mult-view-consistency}
%   \vspace{-0.25in}
% \end{figure}
%
Notably and interestingly, our empirical findings reveal that models pre-trained on MVS and image matching tasks consistently outperform others significantly. This success is attributed to the feature similarity supervision provided by depth annotation \cite{jensen2014large} for MVS and Structure from Motion (SfM) \cite{li2018megadepth} for image matching. Moreover, these tasks yield high-resolution features, emerging as another crucial factor contributing to the superior performance of NSR models.

Furthermore, we observe that 3D-CNN regularization, designed to denoise the 3D cost volume \cite{xu2023iterative, gu2020cascade, cao2022mvsformer}, impedes effective 2D feature learning. Consequently, utilizing features from pre-trained models of image matching tasks without 3D cost volume regularization achieves the best performance and generalization in NSR. Moreover, in our experiments on the EPFL dataset \cite{strecha2008benchmarking}, we make an intriguing observation that patch-wise photometric consistency can be effectively generalized to the feature level, resulting in remarkable improvements. Thus,
 bolstered by patch-wise photometric consistency losses from pre-trained models in both MVSFormer~\cite{cao2022mvsformer} and QuadTree~\cite{tang2022quadtree}
 our approach can produce variations of MVS-NeuS and Match-NeuS. These variations demonstrate state-of-the-art performance on the DTU and EPFL datasets, respectively. 
 \rxl{Additionally, we take a further exploration of the feature priors on grid-based methods such as Neuralangelo~\cite{li2023neuralangelo}, where feature priors still enjoy promising benefits. Thus our method could be extended to different neural representations (MLP or grid-based methods).  }
 \begin{table}   \small
% \footnotesize
\caption{\small A summary of models from various pretext tasks including MIM (masked image modeling), IC (image classification), MDE (monocular depth estimation), SM (stereo matching), SS (semantic segmentation), IM (image matching), and MVS. Only the feature dimensions and resolutions of the highest and lowest features are shown. (Res: Resolution, Dim: Dimension)\label{tab:pretext_tasks}}
\vspace{-0.1in}
\centering
\resizebox{0.65\columnwidth}{!}{
\begin{tabular}{ccccc}
\toprule
\multicolumn{1}{c}{Task} & \multicolumn{1}{c}{Model} & Architect & Feature Res & Feature Dim \\ 
\midrule
\multirow{3}{*}{MIM} & MAE~\cite{he2022masked}& ViT & 1/16  & 1024 \\
& CovMAE~\cite{gao2022convmae} & ViT+CNN   & 1/16,1/4    & 768,256    \\
& CroCo~\cite{weinzaepfel2022croco}  & ViT  & 1/16       & 768         \\
\midrule
\multirow{2}{*}{IC} & Twins~\cite{chu2021twins} & ViT & 1/32, 1/4   & 1024,128    \\
& ConvNeXt~\cite{liu2022convnet} & CNN  & 1/32, 1/4  & 2048,256    \\
\midrule
SS & SegFormer~\cite{xie2021segformer} & ViT & 1/32, 1/4   & 512,64      \\
\midrule
\multirow{2}{*}{MDE} & GLPN~\cite{kim2022global} & ViT  & 1/32, 1/4   & 512,64  \\
& MiDaS~\cite{ranftl2020towards} & ViT & 1/32 & 2048    \\
\midrule
\multirow{2}{*}{SM} & IGEV~\cite{xu2023iterative} & CNN  & 1/32, 1/4   & 160,96      \\
& RAFT-Stereo~\cite{lipson2021raft} & CNN & 1/4 & 256 \\
\midrule
IM & QuadTree~\cite{tang2022quadtree}  & ViT+CNN   & 1/8,1/2     & 256,128     \\
\midrule
\multirow{2}{*}{MVS} & CascadeMVS~\cite{gu2020cascade}  & CNN & 1/8,1 & 32,8  \\
& MVSFormer~\cite{cao2022mvsformer} & ViT+CNN   & 1/8,1  & 32,8    \\
\bottomrule
\end{tabular} }
\vspace{-0.1in}
\end{table}
Our contributions can be highlighted as follows:

\textbf{1)} 
We are the first to conduct a comprehensive evaluation of prior features derived from various pretext tasks on NSR. Additionally, we delve into a detailed comparison of pixel-wise feature consistent loss across these approaches. This comprehensive analysis advances our understanding of the effect of different feature priors for NSR.

\textbf{2)} Our research brings to light a significant finding: models pre-trained on image matching and MVS tasks consistently outperform others in the realm of NSR. This superior performance is attributed to the advantages gained from feature-similarity related pretext training and the utilization of high-resolution features. By identifying these key factors, our work contributes valuable insights that can inform future developments in NSR methods.

\textbf{3)} 
Another notable contribution of our research lies in the successful extension of the patch-wise photometric consistency loss to the feature level. This innovative approach stands as a significant advancement, leading to improved results ~\rxl{regardless of different neural representations (MLP or grid-based methods)}. By bridging the gap between patch-wise and feature-level considerations, our work reaches the state-of-the-art in NSR.
% our work introduces a novel perspective that enhances the state-of-the-art in NSR.

\section{Related Work}
\textbf{Neural Surfaces Reconstruction (NSR).}
Surface reconstruction based on deep learning has garnered significant attention due to its expressive representation capabilities. Representative algorithms focus on utilizing implicit representations to encode the geometry and appearance of 3D scenes through neural networks. These representations include occupancy fields~\cite{mescheder2019occupancy, niemeyer2020differentiable} and SDF~\cite{yariv2020multiview, liu2020dist, zhang2021physg, zhang2021learning, xu2019disn}.
Since the emergence of volume rendering techniques pioneered by NeRF~\cite{mildenhall2021nerf}, numerous NSR methods have seen a dramatic rise~\cite{wang2021neus, yariv2021volume, oechsle2021unisurf, fu2022geo, darmon2022improving, chen2023recovering, guo2022neural, yu2022monosdf,wang2023pet,wang2022hf}. Notably, VolSDF~\cite{yariv2021volume} and NeuS~\cite{wang2021neus} convert volume density to SDF without the need for masks, while UNISURF~\cite{oechsle2021unisurf} unifies both surface and volume rendering.
Concurrently, grid-based representation methods are gaining attention~\cite{muller2022instant, sun2022direct, wang2022go, li2023neuralangelo}, as they offer faster convergence compared to implicit ones with competitive results. Essentially,  note that these explicit NSR methods serve as robust foundations, and our contributions are orthogonal to these approaches.
However, it is crucial to highlight that all these existing methods overlook the importance of multi-view consistency during surface reconstruction. This gap forms the basis for our unique contributions in addressing this aspect.

\noindent \textbf{Photometric Consistency.}
Many previous studies have delved into the application of photometric consistency across various 3D reconstruction tasks~\cite{schonberger2016structure, teed2018deepv2d, vijayanarasimhan2017sfm, yang2018unsupervised, khot2019learning, dai2019mvs2, mur2017orb, yi2023egolocate}. For SfM, several researches aim to optimize camera poses by minimizing photometric errors between corresponding pixels in different images~\cite{schonberger2016structure, teed2018deepv2d, vijayanarasimhan2017sfm}.
Similarly, various self-supervised MVS methods~\cite{yang2018unsupervised, khot2019learning, dai2019mvs2} leverage consistent loss to supervise depth prediction. Moreover, in Simultaneous Localization and Mapping (SLAM), photometric consistency plays a crucial role. By comparing the appearance of projected 3D points in the image with the actual observed images, photometric consistency can detect potential errors or inaccuracies in geometry~\cite{mur2017orb, yi2023egolocate}.

\noindent\textbf{Photometric Consistency in NSR.}
Prior works~\cite{guo2022neural,fu2022geo,darmon2022improving} have incorporated patch-wise photometric consistency directly computed from raw images or pixel-wise feature similarity to enhance NSR training.
NeuralWarp~\cite{guo2022neural} introduces a patch-wise warping loss along with a visibility map that masks incorrect warps arising from occlusion. Similarly, GeoNeus~\cite{fu2022geo} and D-NeuS~\cite{darmon2022improving} focus on locating the zero-level set of geometry network to enforce multi-view geometry constraints, resulting in more accurate reconstruction outcomes.
Despite these advancements, none of these methods thoroughly explore the distinctions between patch-wise and pixel-wise consistent loss, nor do they investigate the potential benefits of incorporating prior features from other pretext visual tasks as opposed to using raw images.

\section{NSR based on Feature Priors}

\begin{figure}[t]
\begin{center}
\includegraphics[width=0.7\linewidth]{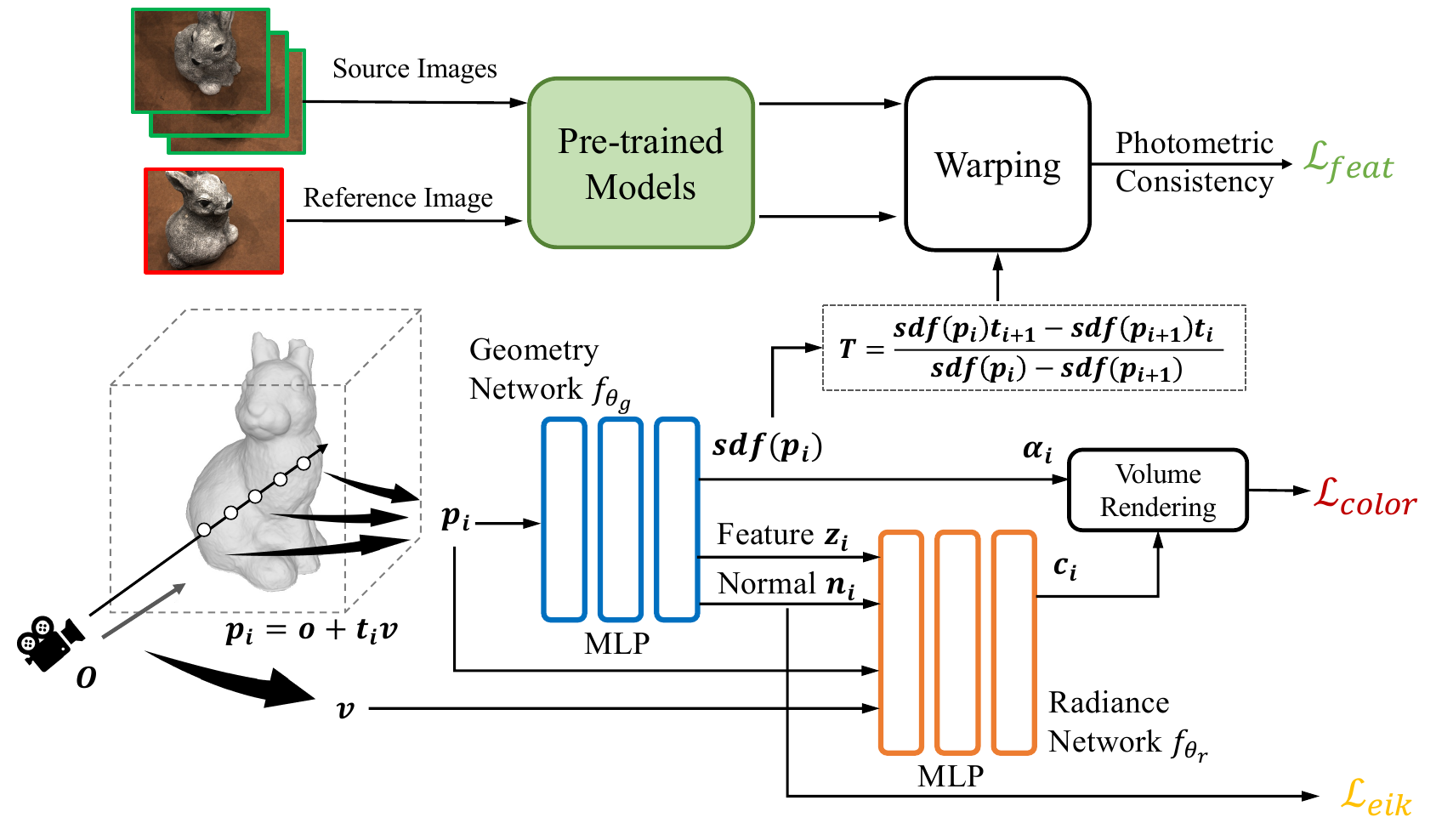}
\vspace{-0.2in}
\end{center}
   \caption{\small Approach overview. Our method is based on NeuS~\cite{wang2021neus}. After achieving surface points from geometry network $f_{\theta_g}$, we further apply the multi-view consistent loss based on features from pre-trained models to improve the reconstruction quality.}
   % We first calculate the surface points (Sec.~\ref{sec:perliminary}) and then leverage the power of pretext task models, whose features can be used for multi-view consistency to significantly improve the reconstruction quality. }
   \label{fig:overview}
\vspace{-0.15in}
\end{figure}

\noindent\textbf{Overview.} 
Given a set of posed multi-view images, the goal of NSR is to reconstruct high-quality surfaces by neural volume rendering, as illustrated in Fig.~\ref{fig:overview}. We revisit the NeuS baseline~\cite{wang2021neus} for multi-view consistency zero-crossing points. We briefly introduce pretext tasks in Sec.~\ref{sec:pretext_tasks} and discuss consistent loss in both pixel-wise and patch-wise terms in Sec.~\ref{sec:consistency}.
By utilizing pre-trained representations from MVS and image matching pretext tasks, we have MVS-NeuS and Match-NeuS, which are compared in the experiments.

\noindent\textbf{NeuS}~\cite{wang2021neus} proposes a novel scheme to achieve unbiased reconstructed surfaces in the first-order approximation of SDF. 
In this scheme, a ray $\{p(t)=o+tv|t>0\}$ is emitted from a camera center at $o$ with view direction $v$.
Colors $\hat{C}(r)$ along a ray can be accumulated by volume rendering~\cite{drebin1988volume}, which is approximated by $N$ discretely sampled points as follows: 
\begin{equation}\label{eq:neus_color_depth}
\hat{C}(r)=\sum_{i=1}^NT_i\alpha_ic_i,\ \ 
% \hat{D}(r)=\sum_{i=1}^N\omega_i\delta_i,
T_i=\prod_{j=1}^{i-1}(1-\alpha_j),
\end{equation}
% \begin{equation}\label{eq:neus_transmittance}
% T_i=\prod_{j=1}^{i-1}(1-\alpha_j), \ \ \omega_i=T_i\alpha_i, \ \ 
% \delta_i =t_{i+1}-t_i,
% \end{equation}
where $\alpha_i$ and $T_i$ are the opacity and accumulated transmittance along the ray respectively.
To achieve the unbiased surface, we can compute $\alpha_i$ by :
% The core formation to achieve the unbiased surface is $\alpha_i$, denoted as:

\begin{equation}\label{eq:neus_alpha}
    \alpha_i=\max(\frac{\Phi_s(sdf(p_i))-\Phi_s(sdf(p_{i+1}))}{\Phi_s(sdf(p_i))},0),
\end{equation}
where $sdf(\cdot)$ is the SDF function modeled by the geometry network $f_{\theta_g}$. $\Phi_s(\cdot)$ is the sigmoid function and $s$ is a learnable parameter. 

\begin{figure}
\vspace{-0.2in}
\begin{center}
\includegraphics[width=0.75\linewidth]{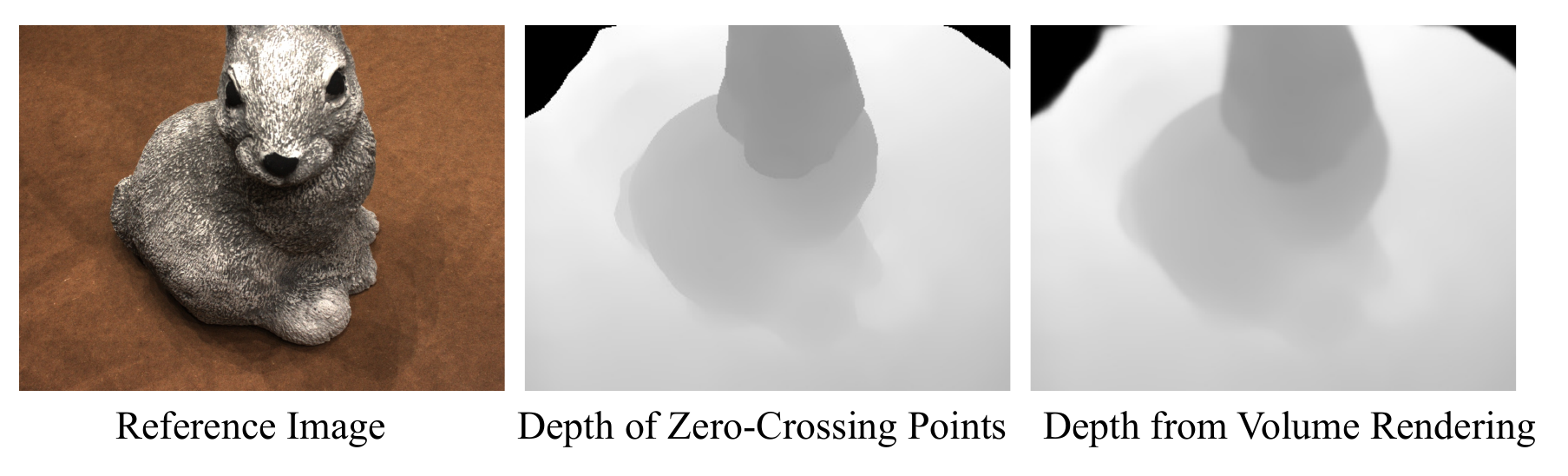}
\vspace{-0.15in}
 \caption{\small Two different ways to locate surface points during the network training at 5000 iterations. Compared with volume rendering, it is more accurate to find the surface points via Eq.~\ref{eq:sdf_zero_level}. Accurate surface points are crucial for applying consistent loss. }
\label{fig:zero_crossing}
\end{center}
\vspace{-0.4in}
\end{figure}

\noindent\textbf{Differentiable Zero-Crossing Points.}
Since achieving photometric consistency involves warping patches or pixels based on a specific depth, we define the depth planes through zero-crossing points achieved from the zero-level set of SDF function as~\cite{darmon2022improving,chen2023recovering}.
So it is crucial to apply photometric consistency specifically to the zero-crossing points of geometry network $f_{\theta_g}$, as shown in Fig.~\ref{fig:zero_crossing}.
We only require the first set of zero-crossing points $T^\ast$ which is visible from the corresponding view. It can be expressed as:
\begin{equation}
\label{eq:sdf_zero_level}
    T^\ast = \left\{t|t=\frac{sdf(p_{t_i})t_{i+1}-sdf(p_{t_{i+1}})t_i}{sdf(p_{t_i})-sdf(p_{t_{i+1}}) }, t_i \in T\right\},
\end{equation}
\begin{equation}
\label{eq:sdf_zero_level2}
     T=\mathop{\arg\min}_i \ \left\{t_i|sdf(p_{t_i})\cdot sdf(p_{t_{t+1}})) <0 \right\}.
\end{equation}

\subsection{Pretext Tasks}
\label{sec:pretext_tasks}

Various models in different pretext visual tasks have been explored as shown in Tab.~\ref{tab:pretext_tasks}. 
Our emphasis is specifically on the features obtained from the feature encoder modules of these pre-trained models. Importantly, many of these models include multi-scale feature maps. In this context, our focus is on the most representative features at both the highest and lowest scales, encompassing high-level semantic understanding and fine-grained low-level details, respectively.

\noindent\textbf{Masked Image Modeling (MIM)}
has been widely investigated recently, as an effective self-supervised task~\cite{he2022masked}.
MIM randomly masks input images at a relatively high rate and subsequently predicts the masked regions based on the context of unmasked areas. 
MIM has gained significant attention recently, with several pioneering works demonstrating its effectiveness in various downstream tasks~\cite{he2022masked, gao2022convmae,weinzaepfel2022croco}. 
In contrast to other self-supervised tasks, 
MIM exhibits favorable fine-grained feature correlations that can potentially benefit NSR learning.

\noindent\textbf{Image Classification (IC)} is a well-established supervised task known for extracting discriminative features. 
In this paper, we investigate models built on CNNs~\cite{liu2022convnet} and ViTs~\cite{chu2021twins} that are pre-trained for image classification on ImageNet~\cite{deng2009imagenet}.
 These models serve as baselines, helping us answer the question of whether engaging in such a classic pretext task also enhances the NSR.

\noindent\textbf{Semantic Segmentation (SS).} Models like~\cite{xie2021segformer, kirillov2023segment} have demonstrated outstanding performance in semantic segmentation by training on large-scale datasets. Unlike classification tasks, these models can capture more detailed, class-specific features, which holds the potential to enhance NSR training.

\noindent\textbf{Monocular Depth Estimation (MDE)} 
aims to predict the depth map of a scene using a single input view, essentially creating a 2D representation of a 3D scene. Several successful works~\cite{ranftl2020towards,kim2022global} have demonstrated the ability to recover fine details of the scene in this process.

\noindent\textbf{Stereo Matching (SM)} is the process of identifying corresponding pixels between two rectified images along epipolar lines, ultimately determining the disparity between these corresponding points. 
 Noteworthy approaches like IGEV~\cite{xu2023iterative} and RAFT-Stereo~\cite{lipson2021raft}  have demonstrated promising results in this task. The learning involved in establishing correspondences bears some similarities to the learning of consistent loss in NSR.

 \noindent\textbf{Multi-View Stereo (MVS)} differs from stereo matching in that it involves multiple input views with known camera poses. However, the MVS problem often faces challenges such as occlusions and varying light conditions.
 Consequently, MVS methods like CasMVSNet~\cite{gu2020cascade} and MVSFormer~\cite{yariv2020multiview} need to enhance feature learning to accurately identify corresponding pixels along epipolar lines.
 It's worth noting that MVS methods may encounter data bias, as discussed in Sec.~\ref{sec:exp_consistent_losses}, particularly on the DTU dataset~\cite{jensen2014large}.

 \noindent\textbf{Image Matching (IM).}
In comparison to stereo matching and MVS, image matching tackles a more intractable problem, \emph{i.e.}, detecting  fine-grained matching pairs between two images without assistance from known camera poses. To solve this challenge, Tand et al.~\cite{tang2022quadtree} utilize a transformer with a lightweight QuadTree attention mechanism to capture long-range dependencies. 
Despite working without constraints from camera poses, we observe that image matching methods can still achieve effective feature representation. 
% even incorporating CNN components.

\subsection{Feature-Level Photometric Consistency}
\label{sec:consistency}

To leverage the consistency of geometry across different views, we utilize the features $\textbf{F}_r$ and $\textbf{F}_s$ extracted from the reference and source images through various pre-trained models respectively. Photometric consistency can be applied to these multi-view features to enforce their alignment and coherence.
Specifically, from the reference viewpoint, we locate a surface point $p$ derived from Eq.~\ref{eq:sdf_zero_level} with its 2D projection coordinate denoted as $x$. This 2D point is then associated with the corresponding point $x'$ in the source images through the homography $H$:
% Specifically, from the reference viewpoint, we can locate a surface point $p$ derived from Eq.~\ref{eq:sdf_zero_level} with its 2D projection coordinate represented as $x$. Subsequently, this 2D point is associated with the corresponding point $x'$ in the source images through the homography $H$:
\begin{equation}
\label{eq:homography}
x'=Hx, \ \  n^Tp +d = 0, \\
\end{equation}
\begin{equation}
     H = K_sR_s(I-\frac{(R_s^{-1} t_s -R_r^{-1}t_r)n^T R_r}{d})R_r^{-1}K_r^{-1},
\end{equation}
where $K_s$ and $K_r$ denote the internal calibration matrix of source and reference cameras, ($R_s$,$R_r$) and ($t_s$,$t_r$) are the $3\times3$ rotation matrix, and 3D translation vector of two views, respectively. $n$ denotes the surface normal, which can be computed by automatic differentiation of the geometry network at each point $p$.
Additionally, $I$ represents the identity matrix and $d$ is the distance from the camera center to the plane of the zero-crossing point (Eq.~\ref{eq:sdf_zero_level}).

\noindent\textbf{Pixel Warping.}
To compute the pixel-level consistency loss, we can evaluate the feature difference between the pixel $x$ in the reference view and its corresponding pixel $x'$ in the source views:
\begin{equation}
    \mathcal{L}_{pixel} = D_{pixel}(\textbf{F}_r(x), \textbf{F}_s(x')),
\end{equation}
where $D_{pixel}$ is the cosine similarity function to measure the feature similarity in pixel-wise.

\noindent\textbf{Patch Warping.} To more effectively incorporate local feature information, we can broaden the concept of pixel warping to patch warping.
To define a small patch $s$ on the surface, traditionally we would compute the surface points using Eq.~\ref{eq:sdf_zero_level} and sample numerous points around those surface points to construct the patch. However, this is computationally expensive. As an alternative, we use the central point $\hat{s}$ of the patch $s$ to represent it and derive the homography matrix $H$. We can perform patch warping from the source view to the reference view and measure the difference between them. The patch-wise consistent loss is 
\begin{equation}
    \mathcal{L}_{patch} = D_{patch}(\textbf{F}_r(\hat{s}), \textbf{F}_s(\hat{s}')),
\end{equation}
where $\hat{s}$, $\hat{s}'$ denote the centers of reference view patches and projected source view patches respectively. $D_{patch}$ can be represented as NCC~\cite{galliani2015massively}, SSIM~\cite{wang2004image}, or patch similarity which measures the average of similarity between corresponding elements in the two patches. \rxl{Since grid-based manners~\cite{li2023neuralangelo} can also locate the surface points via Eq.~\ref{eq:sdf_zero_level}, we easily extend our method to the grid-based methods in Sec.~\ref{sec:grid_based}.} More detailed formulas will be explained in the supplementary.

\noindent\textbf{About Occlusion.} 
Although we can utilize photometric consistency to supervise the geometry network and encourage accurate surface learning, it's important to note that not all surface points are visible in source view images. 
Choosing unsuitable source images would cause prominent degradation to the reconstruction. 
To address this issue, we employ a simple yet effective strategy called Top-k view selection.
Specifically, we sort and select 10 source views based on their relative angles from small to large to the reference perspective. 
Subsequently, we choose Top-k views to compute the patch-wise consistent loss according to the best k photometric scores.
We empirically set k to 4 with proper generalization for most scenes.
However, we have observed that this strategy does not provide significant benefits to pixel-wise consistency due to the incomplete local feature information which results in unstable view selection during training (Sec.~\ref{sec:view_selection}). 

\noindent\textbf{Training Loss.}
\label{sec:training}
To render colors from a specific view, the overall loss function is defined as:
\begin{equation}
    \mathcal{L} = \mathcal{L}_{color} + \lambda_1\cdot\mathcal{L}_{eik} + \lambda_2\cdot\mathcal{L}_{feat},
\end{equation}
where $\mathcal{L}_{color}=\sum_{r\in\mathcal{R}}\Vert \hat{C}(\textbf{r})-C(\textbf{r})\Vert_1$ is the $\ell_1$ loss from minibatch of rays $\mathcal{R}$. $\mathcal{L}_{eik}$ is the eikonal loss defined in~\cite{gropp2020implicit} to regularize the SDF values predicted by the geometry network $f_{\theta_g}$ as depicted in Fig.~\ref{fig:overview}. 

We use feature loss, denoted as $\mathcal{L}_{feat}$, and it is implemented as $\mathcal{L}_{patch}$. Additionally, in Sec.~\ref{sec:exp_pretext_task}, we apply the feature loss as $\mathcal{L}_{pixel}$ to assess the features of pre-trained models derived from different pretext tasks. Our results demonstrate that $\mathcal{L}_{patch}$ exhibits greater robustness compared to $\mathcal{L}_{pixel}$, emphasizing the effectiveness of our patch warping loss.
During the training stage, we set $\lambda_1$ and $\lambda_2$ as 0.1 and 0.5, respectively.

\noindent \textbf{Discussion.} 
While NeuralWarp~\cite{guo2022neural} has introduced a patching-based warping method to transform source view patches to the reference view, it posits that every sampled point along the ray could be projected to other views to obtain the warped colors, leading to the final warped colors through volume rendering. 
In contrast, our approach prioritizes the surface points, offering a more accurate and reasonable method that does not depend on visible maps. Furthermore, we extend the patch warping from RGB pixel-level to feature-level, significantly improving the overall quality of reconstruction.

% \vspace{-2in}
\section{Experiments}

\noindent\textbf{Datasets.} The DTU dataset tested in NSR comprises 15 scenes with 49 or 64 multi-view images collected in an indoor environment with fixed camera poses following~\cite{yariv2020multiview}. The Chamfer Distance (CD) is utilized to evaluate the performance of reconstructed results on the DTU. We also test two outdoor scenes in EPFL dataset~\cite{strecha2008benchmarking}, which has 7 and 11 images at a resolution of $2048\times 3072$, and one of the scenes is particularly challenging due to the occlusions caused by limited viewpoints.

\noindent\textbf{Implementation Details.} We use the same architecture as~\cite{wang2021neus}.
During the training, 512 rays are sampled for each batch following the hierarchical sampling strategy in NeuS~\cite{wang2021neus}. The learning rate is linearly warmed up from 0 to 5e-3 in the first 5k steps and then decayed to 2.5e-5 with a cosine schedule. For each scene, we train the network for 300k steps costing 10 hours on a single A6000.

\subsection{Experiments for Different Pretext Tasks}
\label{sec:exp_pretext_task}
\noindent\textbf{Settings.}
We evaluate the effect of various pretext tasks in Tab.~\ref{tab:pretext_tasks} by applying pixel similarity as the consistency loss on a subset of the DTU dataset including Scan24, Scan37, and Scan106.
These objects are relatively challenging due to difficult lighting conditions and occlusions.
{Because of the quadratic memory footprint of ViT in high-resolution features, we downsample all images into $512\times768$ for a fair comparison.} 
{We only consider the highest and lowest features from the feature encoder as mentioned in Sec.~\ref{sec:pretext_tasks}. 
More experiments on multi-scale features are presented in the supplementary.}

% \begin{figure}[h]
%   \vspace{-0.25in}
%   \centering
%   \includegraphics[width=0.42\columnwidth]{pics/pre-text_models.pdf}
%   \vspace{-0.1in}
%   \caption{\small Quantitative results of using features from different pretext tasks, the dark and light charts represent the results of the model at the highest resolution and lowest resolution, respectively. The red lines indicate the performance of NeuS.}
%   \label{fig:pretext_models}
%   \vspace{-0.25in}
% \end{figure}

\begin{figure}[htb]
\vspace{-0.2in}
    \centering
    \begin{subfigure}{0.3\columnwidth}
        \includegraphics[width=1\columnwidth]{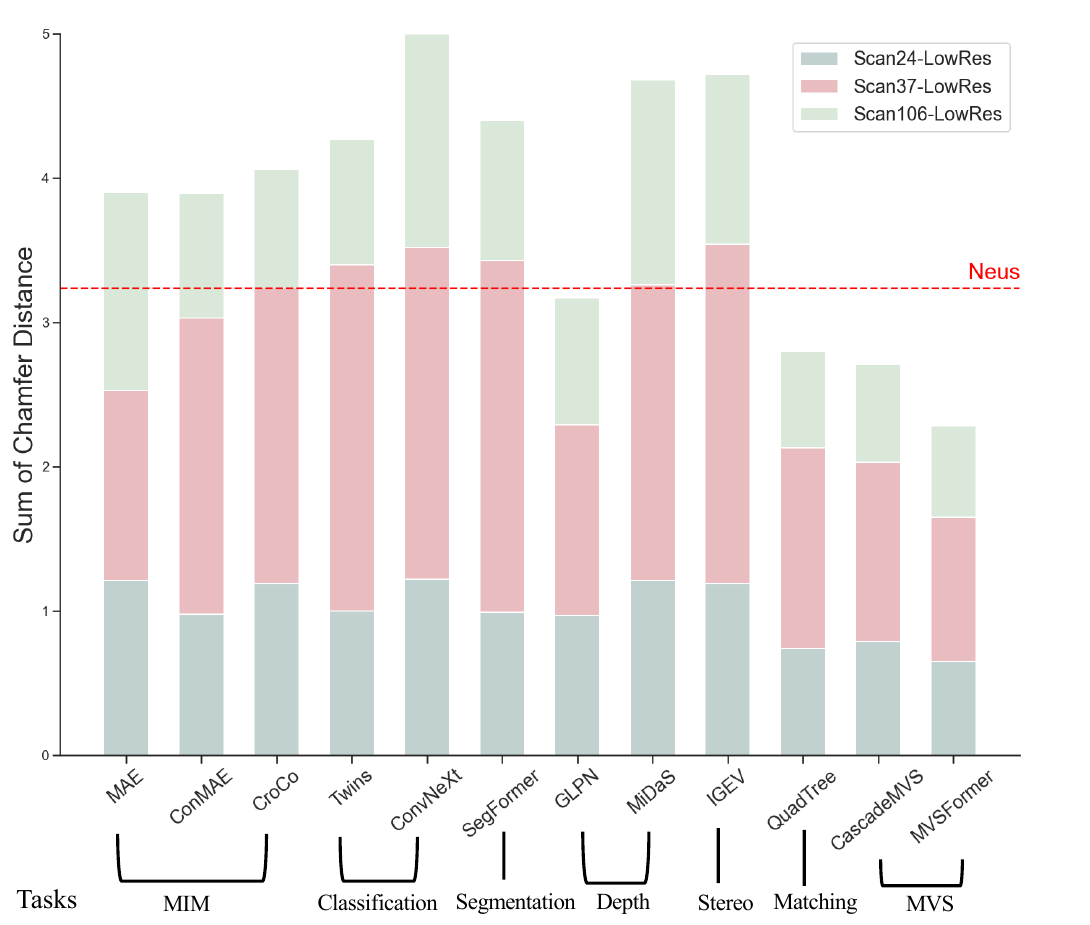}\tabularnewline
        \caption{\small  }\label{fig:pre-text-feature_low}
    \end{subfigure}
    \begin{subfigure}{0.3\columnwidth}
        \includegraphics[width=1\columnwidth]{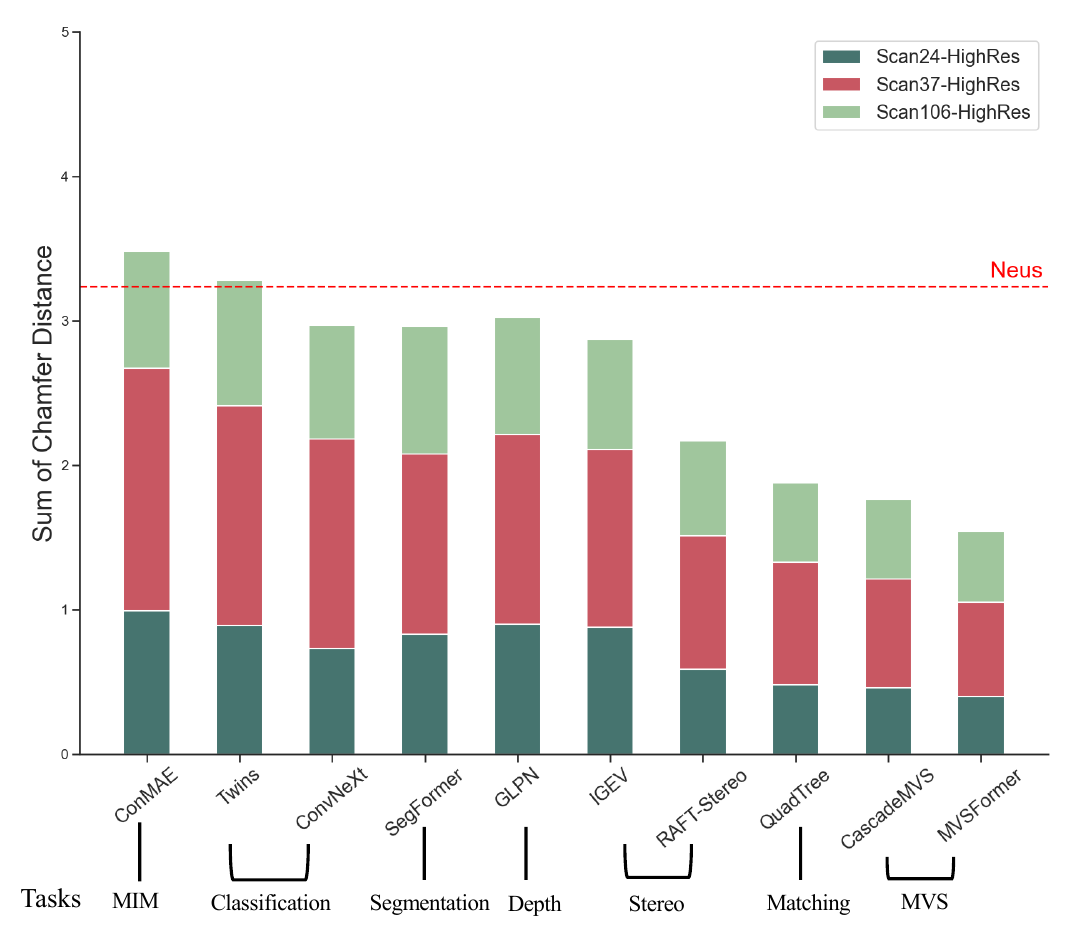}\tabularnewline
        \caption{\small }\label{fig:pre-text-feature_high}
    \end{subfigure}
    \vspace{-0.1in}
    
    % \caption{\small Quantitative results of using features from different pretext tasks. (a) presents the outcomes with the lowest resolution features from different models, while (b) showcases results with the highest resolution feature. The red lines indicate the performance of NeuS, with lower values indicating better performance in terms of Chamfer Distance.}
    \caption{\small Quantitative results of using prior features from different pretext tasks for two scenarios: (a) with the lowest resolution features (high-level features), and (b) with the highest resolution features (low-level features). 
    The red lines indicate the performance of NeuS, with lower values indicating better performance in terms of Chamfer Distance.
    High-resolution features generally outperform the low-resolution ones.}
    \label{fig:pretext_models}
    \vspace{-0.2in}
\end{figure}

\noindent\textbf{In-depth Discussions.}
As depicted in Fig.~\ref{fig:pretext_models}, when employing low-resolution features for the same model, the performance is noticeably worse compared to using high-resolution features. This is because projecting surface points onto low-resolution feature maps does not yield fine-grained effective features, thus hindering the geometry network from accurately learning the correct geometry.
Furthermore, models optimized with high-resolution features from MIM, image classification, semantic segmentation, and monocular depth estimation only perform slightly better than the baseline NeuS~\cite{wang2021neus}. 
For classification~\cite{chu2021twins,liu2022convnet} and segmentation~\cite{xie2021segformer}, we think that these pretext tasks focus on high-level semantic information that is not suitable for NSR.
For MIM~\cite{he2022masked,gao2022convmae,weinzaepfel2022croco}, the most concern is that the effective feature correlations are not fine-grained enough, which limits the performance.
For the depth estimation task~\cite{ranftl2020towards,kim2022global}, we think that it shares little connection with NSR learning.

\begin{table}
\centering
\small
\caption{Quantitative results of CD (mm) on the EPFL dataset with different consistent losses. (Fou: Fountain, Her: Herzjesu)
% % $\star$ indicates models we used for feature priors.  
}
\vspace{-0.1in}
\label{tab:epfl}
% \resizebox{1.0\columnwidth}{!}{
\begin{tabular}{cccccc}
\toprule
\multicolumn{1}{c}{Model} & \multicolumn{1}{c}{Pre-Trained}  & \multicolumn{1}{c}{Consistent Loss} & \multicolumn{1}{c}{Fou} & \multicolumn{1}{c}{Her} & Mean \\ 
\midrule 
\multicolumn{1}{c}{\multirow{4}{*}{Match-NeuS}} & \multicolumn{1}{c}{\multirow{4}{*}{QuadTree~\cite{tang2022quadtree}}}   &   Pixel Similarity & 1.288  & 1.832 & 1.560 \\ 
\multicolumn{1}{c}{}  &   &Patch Similarity & 1.388  & 1.722 & 1.555 \\
\multicolumn{1}{c}{}  &   &Patch NCC & \textbf{1.206}  & \textbf{1.622} & \textbf{1.414} \\
\multicolumn{1}{c}{}  &   &Patch SSIM & 1.218  & 1.678 & 1.448\\ \midrule

\multicolumn{1}{c}{\multirow{4}{*}{MVS-NeuS}} & \multicolumn{1}{c}{\multirow{4}{*}{MVSFormer~\cite{cao2022mvsformer}}}   &   
                           Pixel Similarity & \textbf{1.199}  & 1.698 & 1.449 \\ 
\multicolumn{1}{c}{}  &   &Patch Similarity & \underline{1.212}  & 1.689 & 1.451\\
\multicolumn{1}{c}{}  &   &Patch NCC & 1.242  & \underline{1.646} & \underline{1.444} \\
\multicolumn{1}{c}{}  &   &Patch SSIM & 1.261  & 1.706 & 1.484 \\ \midrule

NeuralWarp~\cite{wang2021neus}  & /  & Patch SSIM& 1.92 & 2.03 & 1.97\\ \midrule

NeuS~\cite{wang2021neus}         &  /  & / & 1.741  & 5.613 & 3.677 \\ \bottomrule
\end{tabular} 
\vspace{-0.15in}
\end{table}

In contrast, models being optimized with feature priors from stereo matching, image matching, and MVS achieve good performance. 
These tasks share an important common factor, \emph{i.e.}, matching feature similarity across different viewpoints, while the supervision is usually annotated by the depth captured from SfM~\cite{li2018megadepth} or specific devices~\cite{jensen2014large}.
Thus, low-level and fine-grained feature correlations should be considered in these manners.  
In general, using feature priors from the same task generally results in a similar performance. 
However, in the stereo matching task, there is a significant performance gap between IGEV~\cite{xu2023iterative} and RAFT-Stereo~\cite{lipson2021raft}, even though these two models have a similar architecture except for the cost volume regularization. 
Specifically, IGEV builds the cost volume as a regularization module that encodes and denoises feature conflicts caused by occlusions and non-Lambertian in the image level through 3D CNNs.
Although 3D cost volume addresses the issues from the 2D feature encoder, we clarify that it could not help to learn essentially better 2D feature representation.
As evidence, features from IGEV achieve inferior performance than RAFT-Stereo, because of the lightweight local cost volume GRU in RAFT-Stereo~\cite{lipson2021raft}, which releases the capacity of 2D features learned by the feature encoder.
From Fig.~\ref{fig:pretext_models}, we observe that QuadTree~\cite{tang2022quadtree} pre-trained with image matching and MVS methods~\cite{gu2020cascade,cao2022mvsformer} enhanced NeuS achieve much better results compared with others. 
As mentioned in Tab.~\ref{tab:pretext_tasks}, models pre-trained on these two tasks enjoy more fine-grained features (1/2 and 1:1).
MVS methods are all pre-trained on the DTU dataset~\cite{jensen2014large}, while the matching method is pre-trained on the MegaDepth~\cite{li2018megadepth}. So MVS manners may suffer from a little data leakage in Fig.~\ref{fig:pretext_models}. 
To be fair, we further compare the performance of image matching and MVS on EPFL dataset in Sec.~\ref{sec:exp_consistent_losses}. 
\begin{figure}[h]
\begin{center}
 \vspace{-0.2in}
\includegraphics[width=0.8\columnwidth]{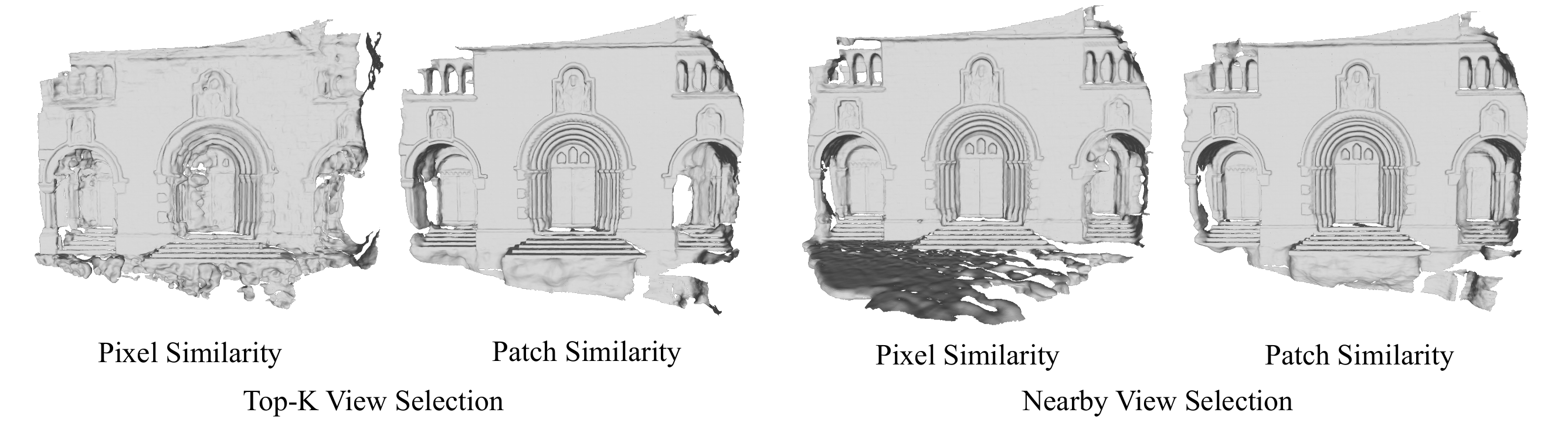}
 \vspace{-0.1in}
 \caption{Qualitative results of different source view selecting strategies based on pixel similarity and patch similarity consistent losses in the Herzjesu of EPFL dataset. \label{fig:candidate_views} }
\end{center}
\vspace{-0.35in}
\end{figure}

\subsection{MVS and Image Matching based NSR}
\label{sec:exp_consistent_losses}
% By incorporating pre-trained representations from MVSFormer~\cite{cao2022mvsformer} and QuadTree~\cite{tang2022quadtree}, 
\rxl{By incorporating features of pre-trained models such as  MVSFormer~\cite{cao2022mvsformer} and QuadTree~\cite{tang2022quadtree}},
our approach can generate variants of MVS-NeuS and Match-NeuS, respectively. Additionally, the evaluation is conducted on EPFL~\cite{strecha2008benchmarking} and includes both pixel-wise and patch-wise losses.

\begin{figure}
\vspace{-0.3in}
\centering
\includegraphics[width=0.72\columnwidth]{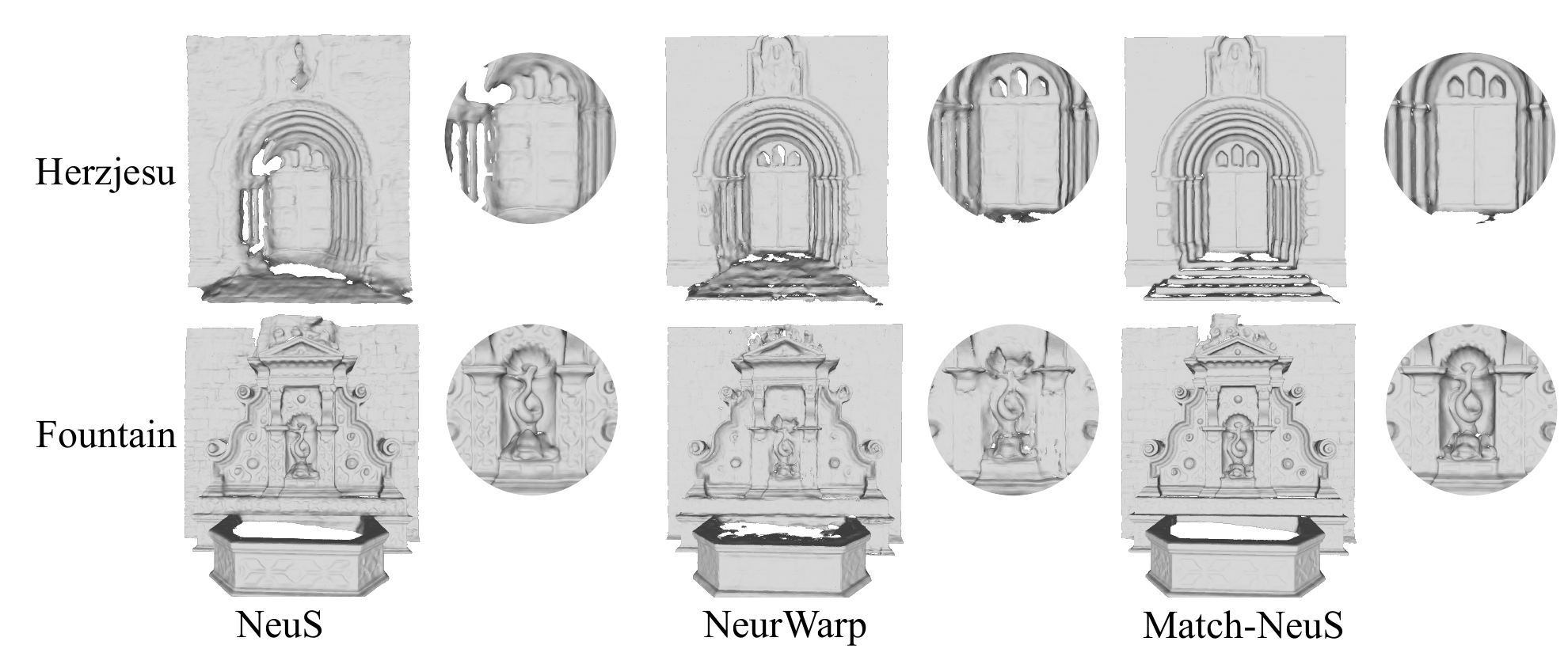}
\vspace{-0.05in}
 \caption{\small Qualitative comparisons on EPFL dataset. \label{fig:exp_epfl} }
\vspace{-0.4in}
\end{figure}

From Tab.~\ref{tab:epfl}, we find that the patch-wise NCC and SSIM could be successfully extended to feature level, and achieve better results compared with pixel similarity for Match-NeuS. Additionally, Match-NeuS with patch NCC outperforms MVS-NeuS, despite features from MVSFormer containing larger resolutions. 
As discussed in Sec.~\ref{sec:exp_pretext_task}, the main reason for the better performance of Match-NeuS is that image matching models are pre-trained without the cost volume regularization, because no camera poses could be used to formulate the 3D cost volume. 
Furthermore, due to the prohibitive computation for the 3D CNN in the cost volume regularization, MVS methods should reduce the 2D feature channels for efficiency. 
Thus the matching-based QuadTree has 256 feature channels, while MVSformer only contains 8 feature channels with less capacity.
In Fig.~\ref{fig:exp_epfl}, we show that Match-NeuS with patch NCC recovers more detailed structures in Fountain, while reconstructing smoother walls.

\begin{table} 
\small
\vspace{-0.1in}
\caption{\small Ablation study of CD (mm) on matching features \textbf{w/wo} attention modules.
We further compare them to the results of high-resolution features (1/2) without attention modules.
}
\vspace{-0.1in}
\centering
\label{tab:epfl_transformer}
% \resizebox{0.95\columnwidth}{!}{
\begin{tabular}{lccccc}
\toprule
\multicolumn{1}{c}{Model}          & Consistent Loss     & Res.  & Fountain & Herzjesu & Mean \\
\midrule 
Match-NeuS &   Pixel Similarity & $1/2$ & \textbf{1.288}  & \textbf{1.832} & \textbf{1.560}  \\
Match-NeuS       & Pixel Similarity &  $1/8$ & 1.634     & 3.360     & 2.497 \\
\quad + Attention & Pixel Similarity & $1/8$ & 1.502      & 3.010        & 2.256 \\
\bottomrule
\end{tabular}
\vspace{-0.2in}
\end{table}

\noindent\textbf{View Selection.}
\label{sec:view_selection}
We compare the view selection for both pixel-wise and patch-wise losses in Fig.~\ref{fig:candidate_views},
From Fig.~\ref{fig:candidate_views}, the Top-k view selection mentioned in Sec.~\ref{sec:consistency} fails to adaptively choose reasonable source views through pixel-wise feature similarity, resulting in poor reconstruction.
Thus we need to manually select nearby source views to stabilize the training of pixel-wise feature similarity.
However, the Top-k view selection based on patch-wise feature similarity can achieve stable training and better reconstruction. 
Because local patches usually capture more information with larger perceptive fields, effectively addressing occlusion issues.

\noindent\textbf{Features with Attention Modules.} We observe that QuadTree employs the interleaved attention mechanism~\cite{tang2022quadtree} including self and cross-attention modules. Attention modules enable explicit feature matching across different views and enhance the feature capacity. 
However, due to its high computational cost, the transformer layer is designed to refine only 1/8 feature scale. 
Hence this improvement fails to be transferred to high-resolution feature learning (1/2).
We have presented a comparison in Tab.~\ref{tab:epfl_transformer} to showcase this improvement. 
It is evident that by incorporating the features after attention learning, Match-NeuS improves its performance, but it still performs worse than high-resolution feature learning.

\subsection{Extension of Feature Prior on Grid-based Representation.}
\label{sec:grid_based}
\begin{figure}[htbp]
\vspace{-0.25in}
\centering
\includegraphics[width=0.65\columnwidth]{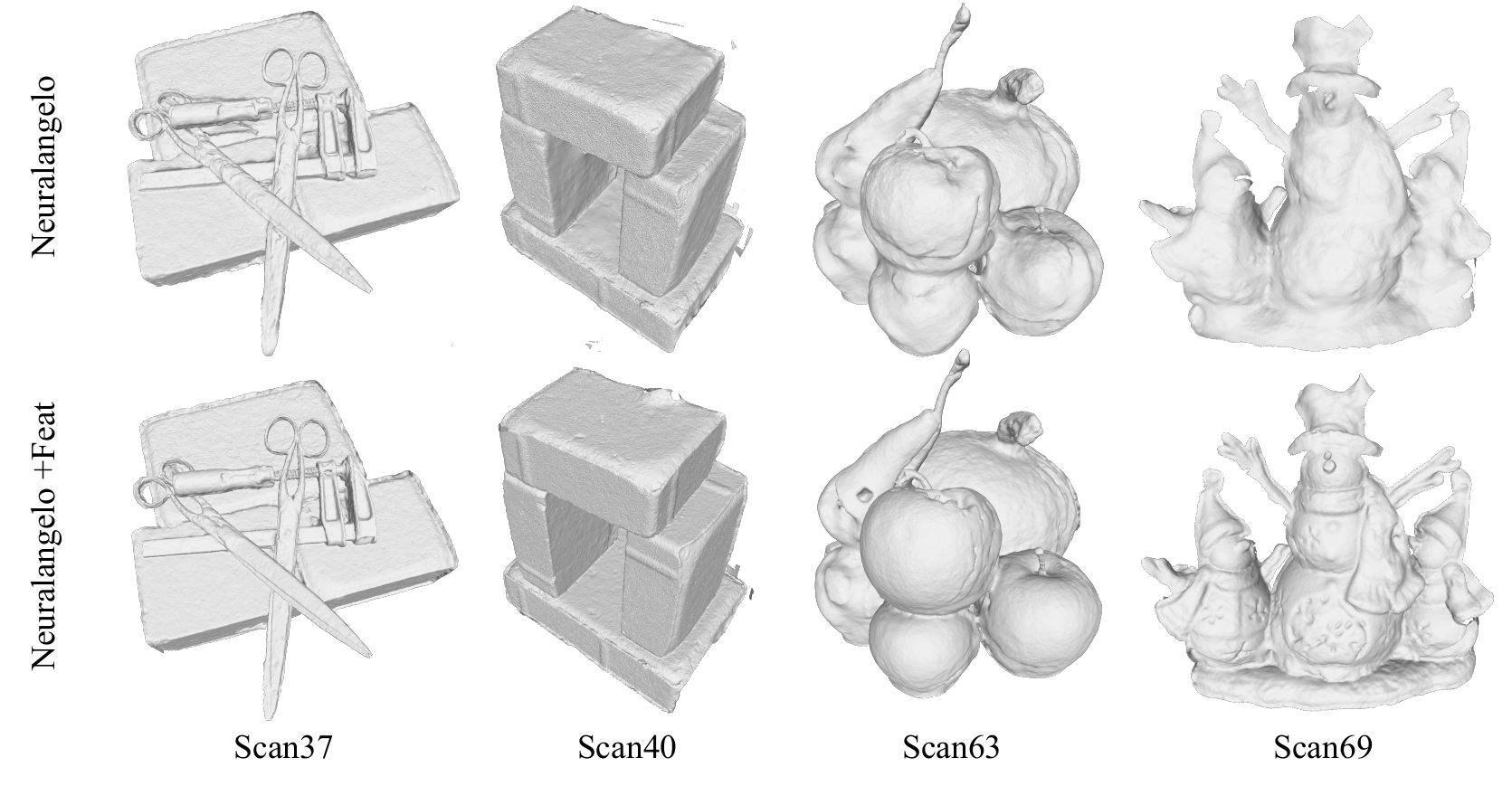}
\caption{Qualitative results of Neuralangelo with feature priors.}
\label{fig:neuralango}
\vspace{-0.25in}
\end{figure}
We incorporate feature priors into grid-based Neuralangelo~\cite{li2023neuralangelo} as in Fig~\ref{fig:neuralango}. Specifically, we train Neuralangelo with MVSFormer feature priors on DTU dataset for 300k iterations. The inclusion of MVS features enhances the accuracy of geometry and accelerates convergence, especially in intricate objects like scissors' handles in Scan37 and the snowman in Scan69. Moreover, Neuralangelo with feature priors achieves smoother flat surfaces in Scan40 and Scan63.  

% We find feature priors are still beneficial for NSR training, regardless of different neural representations (MLP or Grid-based methods). 
% Grid-based methods~\cite{muller2022instant, sun2022direct, wang2022go, li2023neuralangelo} show their faster convergence and generating fine-grained details to implicit methods, We extend our feature priors to one of the pioneering works such as Neuralangelo~\cite{li2023neuralangelo},  We train Neuralangelo with MVSFormer feature priors on Scan37 of the DTU dataset and Barn of Tanks\&Temples dataset for 200k iterations. The 
% inclusion of MVS features results in faster convergence and more accurate geometry, particularly in objects like scissors’ handle of Scan37 and railings of Barn. Neuralangelo with feature prior enhances the smoothness of flat planes, and captures intricate details as shown in Fig.~\ref{fig:neuralango}.

\begin{figure*}[t]
\begin{center}
\includegraphics[width=1.0\columnwidth]{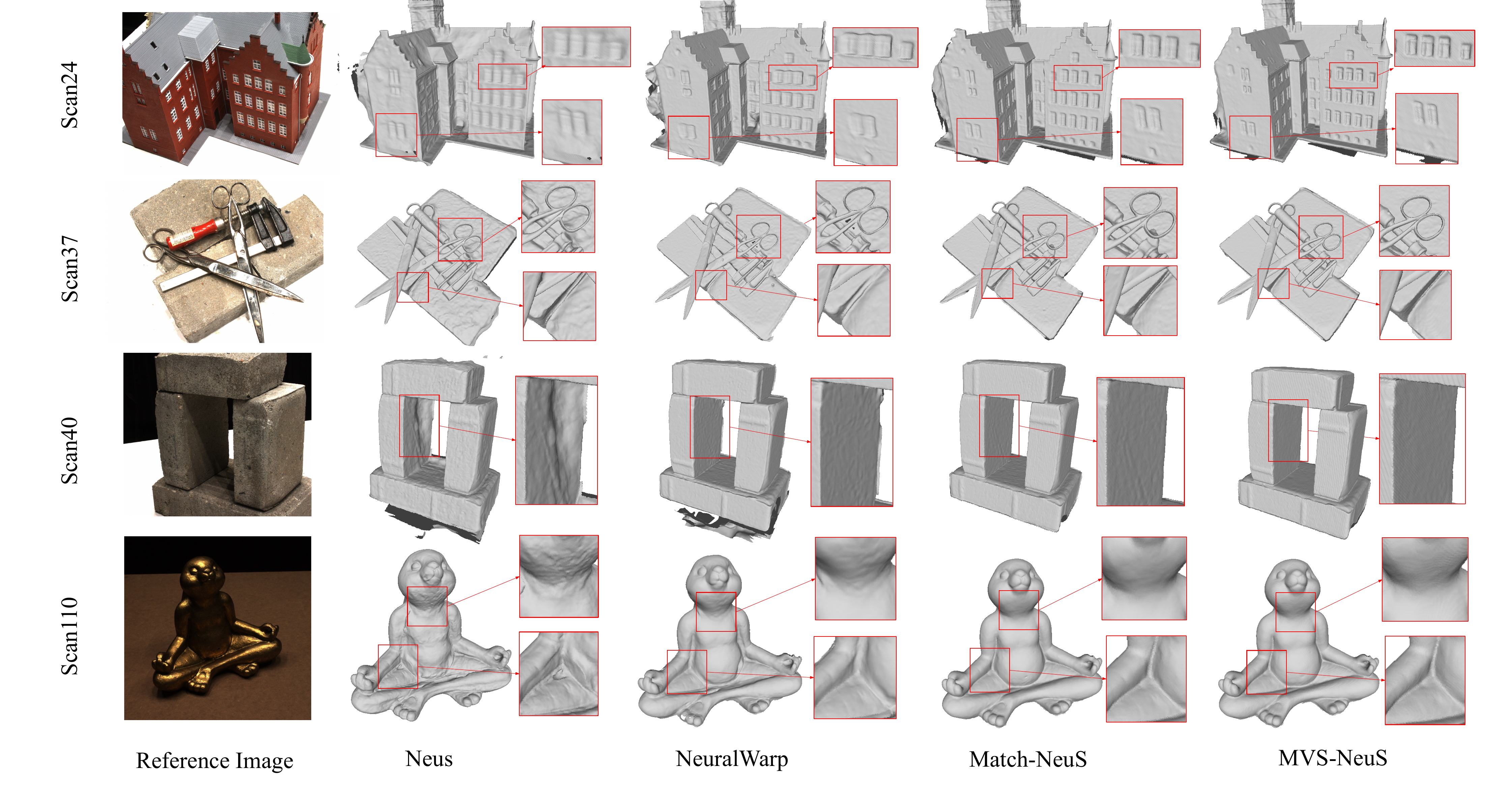}
\vspace{-0.15in}
   \caption{\small Comparisons of surface reconstruction in DTU dataset.}
   \label{fig:dtu}
\end{center}
\vspace{-0.15in}
\end{figure*}

\subsection{Full Models Compared on DTU}
\begin{table*} \small
\vspace{-0.1in}
\centering
\caption{\small Results of CD (mm) compared on DTU dataset (lower is better). The upper part methods enjoy additional annotations such as masks (IDR and COLMAP) or custom filtering (MVSDF). Note that the results of GeoNeuS* are re-trained by ourselves, 
The best results are in \textbf{bold}.\label{tab:dtu_result}}
\vspace{-0.15in}
 \resizebox{1.0\linewidth}{!}{
\begin{tabular}{ccccccccccccccccc}
\toprule 
Scan & 24 & 37 & 40 & 55 & 63 & 65 & 69 & 83 & 97 & 105 & 106 & 110 & 114 & 118 & 122 & Mean\tabularnewline
\midrule 
COLMAP~\cite{schonberger2016pixelwise} & 0.45 & 0.91 & 0.37 & 0.37 & 0.90 & 1.00 & 0.54 & 1.22 & 1.08 & 0.64 & 0.48 & 0.59 & 0.32 & 0.45 & 0.43 & 0.65\tabularnewline
MVSDF~\cite{zhang2021learning} & 0.93 & 1.75 & 0.88 & 0.44 & 1.11 & 0.90 & 0.75 & 1.26 & 1.02 & 1.35 & 0.87 & 0.84 & 0.34 & 0.47 & 0.47 & 0.88\tabularnewline
IDR~\cite{yariv2020multiview} & 1.63 & 1.87 & 0.63 & 0.48 & 1.04 & 0.79 & 0.77 & 1.33 & 1.16 & 0.76 & 0.67 & 0.90 & 0.42 & 0.51 & 0.53 & 0.90\tabularnewline
\midrule 
NeRF~\cite{mildenhall2021nerf} & 1.90 & 1.60 & 1.85 & 0.58 & 2.28 & 1.27 & 1.47 & 1.67 & 2.05 & 1.07 & 0.88 & 2.53 & 1.06 & 1.15 & 0.96 & 1.49\tabularnewline
NeuS~\cite{wang2021neus} & 1.37 & 1.21 & 0.73 & 0.40 & 1.20 & 0.70 & 0.72 & \textbf{1.01} & 1.16 & 0.82 & 0.66 & 1.69 & 0.39 & 0.49 & 0.51 & 0.87\tabularnewline
VolSDF~\cite{yariv2021volume} & 1.14 & 1.26 & 0.81 & 0.49 & 1.25 & 0.70 & 0.72 & 1.29 & 1.18 & 0.70 & 0.66 & 1.08 & 0.42 & 0.61 & 0.55 & 0.86\tabularnewline
NeuralWarp~\cite{darmon2022improving} & 0.49 & 0.71 & 0.38 & 0.38 & \textbf{0.79} & 0.81 & 0.82 & 1.20 & 1.06 & 0.68 & 0.66 & 0.74 & 0.41 & 0.63 & 0.51 & 0.68\tabularnewline
GeoNeuS{*}~\cite{fu2022geo} & 0.40 & 0.69 & 0.36 & 0.37 & \textbf{0.79} & {0.78} & \textbf{0.48} & 1.17 & 0.88 & \textbf{0.60} & 0.50 & 0.73 & 0.30 & 0.39 & 0.41 & 0.59 \tabularnewline
D-NeuS~\cite{chen2023recovering} & 0.44 & 0.79 & 0.35 & 0.39 & 0.88 & 0.58 & 0.55 & 1.35 & 0.91 & 0.76 & \textbf{0.40} & 0.72 & 0.31 & 0.39 & \textbf{0.39} & 0.61\tabularnewline
PET-NeuS~\cite{wang2023pet} & 0.56 & 0.75 & 0.68 & 0.36 & 0.87 & 0.76 & 0.69 & 1.33 & 1.08 & 0.66 & 0.51 & 1.04 & 0.34 & 0.51 & 0.48 & 0.71 \tabularnewline
\midrule
MVS-NeuS  & \textbf{0.37} & \textbf{0.63} &  \textbf{0.31} & \textbf{0.32} & \textbf{0.79} & \textbf{0.51} & 0.50 & 1.12 & \textbf{0.87} & 0.62 & 0.46 & \textbf{0.64} & \textbf{0.29} & \textbf{0.38} & 0.41 & \textbf{0.548}\tabularnewline
Match-NeuS  & 0.39  &0.67 &0.32  &0.33  &\textbf{0.79}  &\textbf{0.51}  &0.49  & 1.21 & \textbf{0.87}  & 0.63  & 0.48 & 0.71  & \textbf{0.29} & \textbf{0.38} & 0.41 & 0.565\tabularnewline
\bottomrule 
\end{tabular}
}
\vspace{-0.2in}
\end{table*}

We compare MVS-NeuS and Match-NeuS with patch NCC as the consistent loss to other competitors on 15 scenes in the DTU dataset. 
Tab.~\ref{tab:dtu_result} shows the quantitative results. Notably, our methods outperform other competitors by a large margin. 
Furthermore, we observed the qualitative results in Fig.~\ref{fig:dtu}
% that our methods achieve good performance in both smooth flat plane and abrupt depth regions. 
that our method can accurately reconstruct the geometry of these scenes.
Specifically, our method is capable of not only modeling smooth flat planes but also clearly capturing gaps between two planes or the handle of scissors. 
In contrast, many other methods tend to fuse together in Scan37 with obvious artifacts. 
Quantitative results in Tab.~\ref{tab:dtu_result} show that our model significantly outperforms other competing approaches.
% our model could generate more accurate geometry.

% Our quantitative comparison of all methods on the DTU dataset is presented in Table~\ref{tab:dtu_result}. Most results are cited from~\cite{chen2023recovering}, except for GeoNeuS, which we retrained. 
% Our approach, which employed the consistent loss using features from MVSFormer and QuadTree, outperformed NeuS by 50\%, as shown in Table~\ref{tab:dtu_result}. 
% Furthermore, we observed in Fig~\ref{fig:dtu} that the plane became flatter in Scan37 and Scan24 after applying the photometric consistency loss.

\section{Conclusion}
In this paper, we conducted an evaluation of various pretext tasks and consistent losses for NSR.
Our findings showed that the feature priors from some pretext visual tasks including MVS and image matching can significantly improve the NSR performance.
We also empirically observed that features with higher resolution consistently outperform the lower ones.
Additionally, we further explored optimizing MVS and image matching enhanced NSR with both pixel-wise and patch-wise consistent losses. The results showed that the patch-wise consistent loss could be successfully generalized to the feature level and achieve superior reconstruction for the matching-based Match-NeuS.

\section*{Acknowledgements}
This work was supported in part by the Shanghai Technology Development and Entrepreneurship Platform for Neuromorphic and AI SoC, the National Natural Science
Foundation of China under Grant 62076067 and Shanghai Municipal Science and Technology Major Project (NO. 2021SHZDZX0103).
This work was performed using the CFFF platform of Fudan University. Yanwei Fu is with School of Data Science,  Fudan ISTBI—ZJNU
Algorithm Centre for Brain-inspired Intelligence, Fudan University,  Shanghai Key Lab of
Intelligent Information Processing, and Technology Innovation
Center of Calligraphy and Painting Digital Generation, Ministry
of Culture and Tourism, China. Email: yanweifu@fudan.edu.cn

% \clearpage  % TODO REVIEW/FINAL: This \clearpage needs to be removed from both review and camera-ready versions.

% ---- Bibliography ----
%
% BibTeX users should specify bibliography style 'splncs04'.
% References will then be sorted and formatted in the correct style.
%
\bibliographystyle{splncs04}
\bibliography{main}

\begin{thebibliography}{10}
\providecommand{\url}[1]{\texttt{#1}}
\providecommand{\urlprefix}{URL }
\providecommand{\doi}[1]{https://doi.org/#1}

\bibitem{barnes2009patchmatch}
Barnes, C., Shechtman, E., Finkelstein, A., Goldman, D.B.: Patchmatch: A randomized correspondence algorithm for structural image editing. ACM Trans. Graph.  \textbf{28}(3), ~24 (2009)

\bibitem{cao2022mvsformer}
Cao, C., Ren, X., Fu, Y.: Mvsformer: Multi-view stereo by learning robust image features and temperature-based depth. Transactions of Machine Learning Research

\bibitem{chen2023recovering}
Chen, D., Zhang, P., Feldmann, I., Schreer, O., Eisert, P.: Recovering fine details for neural implicit surface reconstruction. In: Proceedings of the IEEE/CVF Winter Conference on Applications of Computer Vision. pp. 4330--4339 (2023)

\bibitem{chu2021twins}
Chu, X., Tian, Z., Wang, Y., Zhang, B., Ren, H., Wei, X., Xia, H., Shen, C.: Twins: Revisiting the design of spatial attention in vision transformers. Advances in Neural Information Processing Systems  \textbf{34},  9355--9366 (2021)

\bibitem{dai2019mvs2}
Dai, Y., Zhu, Z., Rao, Z., Li, B.: Mvs2: Deep unsupervised multi-view stereo with multi-view symmetry. In: 2019 International Conference on 3D Vision (3DV). pp.~1--8. Ieee (2019)

\bibitem{darmon2022improving}
Darmon, F., Bascle, B., Devaux, J.C., Monasse, P., Aubry, M.: Improving neural implicit surfaces geometrym fan with patch warping. In: Proceedings of the IEEE/CVF Conference on Computer Vision and Pattern Recognition. pp. 6260--6269 (2022)

\bibitem{deng2009imagenet}
Deng, J., Dong, W., Socher, R., Li, L.J., Li, K., Fei-Fei, L.: Imagenet: A large-scale hierarchical image database. In: 2009 IEEE conference on computer vision and pattern recognition. pp. 248--255. Ieee (2009)

\bibitem{drebin1988volume}
Drebin, R.A., Carpenter, L., Hanrahan, P.: Volume rendering. ACM Siggraph Computer Graphics  \textbf{22}(4),  65--74 (1988)

\bibitem{fu2022geo}
Fu, Q., Xu, Q., Ong, Y.S., Tao, W.: Geo-neus: geometry-consistent neural implicit surfaces learning for multi-view reconstruction. arXiv preprint arXiv:2205.15848  (2022)

\bibitem{galliani2015massively}
Galliani, S., Lasinger, K., Schindler, K.: Massively parallel multiview stereopsis by surface normal diffusion. In: Proceedings of the IEEE International Conference on Computer Vision. pp. 873--881 (2015)

\bibitem{gao2022convmae}
Gao, P., Ma, T., Li, H., Dai, J., Qiao, Y.: Convmae: Masked convolution meets masked autoencoders. arXiv preprint arXiv:2205.03892  (2022)

\bibitem{gropp2020implicit}
Gropp, A., Yariv, L., Haim, N., Atzmon, M., Lipman, Y.: Implicit geometric regularization for learning shapes. arXiv preprint arXiv:2002.10099  (2020)

\bibitem{gu2020cascade}
Gu, X., Fan, Z., Zhu, S., Dai, Z., Tan, F., Tan, P.: Cascade cost volume for high-resolution multi-view stereo and stereo matching. In: Proceedings of the IEEE/CVF conference on computer vision and pattern recognition. pp. 2495--2504 (2020)

\bibitem{guo2022neural}
Guo, H., Peng, S., Lin, H., Wang, Q., Zhang, G., Bao, H., Zhou, X.: Neural 3d scene reconstruction with the manhattan-world assumption. In: Proceedings of the IEEE/CVF Conference on Computer Vision and Pattern Recognition. pp. 5511--5520 (2022)

\bibitem{he2022masked}
He, K., Chen, X., Xie, S., Li, Y., Doll{\'a}r, P., Girshick, R.: Masked autoencoders are scalable vision learners. In: Proceedings of the IEEE/CVF Conference on Computer Vision and Pattern Recognition. pp. 16000--16009 (2022)

\bibitem{jensen2014large}
Jensen, R., Dahl, A., Vogiatzis, G., Tola, E., Aan{\ae}s, H.: Large scale multi-view stereopsis evaluation. In: Proceedings of the IEEE conference on computer vision and pattern recognition. pp. 406--413 (2014)

\bibitem{kazhdan2013screened}
Kazhdan, M., Hoppe, H.: Screened poisson surface reconstruction. ACM Transactions on Graphics (ToG)  \textbf{32}(3),  1--13 (2013)

\bibitem{khot2019learning}
Khot, T., Agrawal, S., Tulsiani, S., Mertz, C., Lucey, S., Hebert, M.: Learning unsupervised multi-view stereopsis via robust photometric consistency. arXiv preprint arXiv:1905.02706  (2019)

\bibitem{kim2022global}
Kim, D., Ka, W., Ahn, P., Joo, D., Chun, S., Kim, J.: Global-local path networks for monocular depth estimation with vertical cutdepth. arXiv preprint arXiv:2201.07436  (2022)

\bibitem{kirillov2023segment}
Kirillov, A., Mintun, E., Ravi, N., Mao, H., Rolland, C., Gustafson, L., Xiao, T., Whitehead, S., Berg, A.C., Lo, W.Y., et~al.: Segment anything. arXiv preprint arXiv:2304.02643  (2023)

\bibitem{labatut2007efficient}
Labatut, P., Pons, J.P., Keriven, R.: Efficient multi-view reconstruction of large-scale scenes using interest points, delaunay triangulation and graph cuts. In: 2007 IEEE 11th international conference on computer vision. pp.~1--8. IEEE (2007)

\bibitem{li2023neuralangelo}
Li, Z., M{\"u}ller, T., Evans, A., Taylor, R.H., Unberath, M., Liu, M.Y., Lin, C.H.: Neuralangelo: High-fidelity neural surface reconstruction. In: Proceedings of the IEEE/CVF Conference on Computer Vision and Pattern Recognition. pp. 8456--8465 (2023)

\bibitem{li2018megadepth}
Li, Z., Snavely, N.: Megadepth: Learning single-view depth prediction from internet photos. In: Proceedings of the IEEE conference on computer vision and pattern recognition. pp. 2041--2050 (2018)

\bibitem{lipson2021raft}
Lipson, L., Teed, Z., Deng, J.: Raft-stereo: Multilevel recurrent field transforms for stereo matching. In: 2021 International Conference on 3D Vision (3DV). pp. 218--227. IEEE (2021)

\bibitem{liu2020dist}
Liu, S., Zhang, Y., Peng, S., Shi, B., Pollefeys, M., Cui, Z.: Dist: Rendering deep implicit signed distance function with differentiable sphere tracing. In: Proceedings of the IEEE/CVF Conference on Computer Vision and Pattern Recognition. pp. 2019--2028 (2020)

\bibitem{liu2022convnet}
Liu, Z., Mao, H., Wu, C.Y., Feichtenhofer, C., Darrell, T., Xie, S.: A convnet for the 2020s. Proceedings of the IEEE/CVF Conference on Computer Vision and Pattern Recognition (CVPR)  (2022)

\bibitem{mescheder2019occupancy}
Mescheder, L., Oechsle, M., Niemeyer, M., Nowozin, S., Geiger, A.: Occupancy networks: Learning 3d reconstruction in function space. In: Proceedings of the IEEE/CVF conference on computer vision and pattern recognition. pp. 4460--4470 (2019)

\bibitem{mildenhall2021nerf}
Mildenhall, B., Srinivasan, P.P., Tancik, M., Barron, J.T., Ramamoorthi, R., Ng, R.: Nerf: Representing scenes as neural radiance fields for view synthesis. Communications of the ACM  \textbf{65}(1),  99--106 (2021)

\bibitem{muller2022instant}
M{\"u}ller, T., Evans, A., Schied, C., Keller, A.: Instant neural graphics primitives with a multiresolution hash encoding. ACM Transactions on Graphics (ToG)  \textbf{41}(4),  1--15 (2022)

\bibitem{mur2017orb}
Mur-Artal, R., Tard{\'o}s, J.D.: Orb-slam2: An open-source slam system for monocular, stereo, and rgb-d cameras. IEEE transactions on robotics  \textbf{33}(5),  1255--1262 (2017)

\bibitem{niemeyer2020differentiable}
Niemeyer, M., Mescheder, L., Oechsle, M., Geiger, A.: Differentiable volumetric rendering: Learning implicit 3d representations without 3d supervision. In: Proceedings of the IEEE/CVF Conference on Computer Vision and Pattern Recognition. pp. 3504--3515 (2020)

\bibitem{oechsle2021unisurf}
Oechsle, M., Peng, S., Geiger, A.: Unisurf: Unifying neural implicit surfaces and radiance fields for multi-view reconstruction. In: Proceedings of the IEEE/CVF International Conference on Computer Vision. pp. 5589--5599 (2021)

\bibitem{ranftl2020towards}
Ranftl, R., Lasinger, K., Hafner, D., Schindler, K., Koltun, V.: Towards robust monocular depth estimation: Mixing datasets for zero-shot cross-dataset transfer. IEEE transactions on pattern analysis and machine intelligence  \textbf{44}(3),  1623--1637 (2020)

\bibitem{schonberger2016structure}
Schonberger, J.L., Frahm, J.M.: Structure-from-motion revisited. In: Proceedings of the IEEE conference on computer vision and pattern recognition. pp. 4104--4113 (2016)

\bibitem{schonberger2016pixelwise}
Sch{\"o}nberger, J.L., Zheng, E., Frahm, J.M., Pollefeys, M.: Pixelwise view selection for unstructured multi-view stereo. In: Computer Vision--ECCV 2016: 14th European Conference, Amsterdam, The Netherlands, October 11-14, 2016, Proceedings, Part III 14. pp. 501--518. Springer (2016)

\bibitem{strecha2008benchmarking}
Strecha, C., Von~Hansen, W., Van~Gool, L., Fua, P., Thoennessen, U.: On benchmarking camera calibration and multi-view stereo for high resolution imagery. In: 2008 IEEE conference on computer vision and pattern recognition. pp.~1--8. Ieee (2008)

\bibitem{sun2022direct}
Sun, C., Sun, M., Chen, H.T.: Direct voxel grid optimization: Super-fast convergence for radiance fields reconstruction. In: Proceedings of the IEEE/CVF Conference on Computer Vision and Pattern Recognition. pp. 5459--5469 (2022)

\bibitem{tang2022quadtree}
Tang, S., Zhang, J., Zhu, S., Tan, P.: Quadtree attention for vision transformers. ICLR  (2022)

\bibitem{teed2018deepv2d}
Teed, Z., Deng, J.: Deepv2d: Video to depth with differentiable structure from motion. arXiv preprint arXiv:1812.04605  (2018)

\bibitem{vijayanarasimhan2017sfm}
Vijayanarasimhan, S., Ricco, S., Schmid, C., Sukthankar, R., Fragkiadaki, K.: Sfm-net: Learning of structure and motion from video. arXiv preprint arXiv:1704.07804  (2017)

\bibitem{wang2022go}
Wang, J., Bleja, T., Agapito, L.: Go-surf: Neural feature grid optimization for fast, high-fidelity rgb-d surface reconstruction. arXiv preprint arXiv:2206.14735  (2022)

\bibitem{wang2021neus}
Wang, P., Liu, L., Liu, Y., Theobalt, C., Komura, T., Wang, W.: Neus: Learning neural implicit surfaces by volume rendering for multi-view reconstruction. arXiv preprint arXiv:2106.10689  (2021)

\bibitem{wang2022hf}
Wang, Y., Skorokhodov, I., Wonka, P.: Hf-neus: Improved surface reconstruction using high-frequency details. Advances in Neural Information Processing Systems  \textbf{35},  1966--1978 (2022)

\bibitem{wang2023pet}
Wang, Y., Skorokhodov, I., Wonka, P.: Pet-neus: Positional encoding tri-planes for neural surfaces. In: Proceedings of the IEEE/CVF Conference on Computer Vision and Pattern Recognition. pp. 12598--12607 (2023)

\bibitem{wang2004image}
Wang, Z., Bovik, A.C., Sheikh, H.R., Simoncelli, E.P.: Image quality assessment: from error visibility to structural similarity. IEEE transactions on image processing  \textbf{13}(4),  600--612 (2004)

\bibitem{weinzaepfel2022croco}
Weinzaepfel, P., Leroy, V., Lucas, T., Br{\'e}gier, R., Cabon, Y., Arora, V., Antsfeld, L., Chidlovskii, B., Csurka, G., Revaud, J.: Croco: Self-supervised pre-training for 3d vision tasks by cross-view completion. arXiv preprint arXiv:2210.10716  (2022)

\bibitem{xie2021segformer}
Xie, E., Wang, W., Yu, Z., Anandkumar, A., Alvarez, J.M., Luo, P.: Segformer: Simple and efficient design for semantic segmentation with transformers. Advances in Neural Information Processing Systems  \textbf{34},  12077--12090 (2021)

\bibitem{xu2023iterative}
Xu, G., Wang, X., Ding, X., Yang, X.: Iterative geometry encoding volume for stereo matching. arXiv preprint arXiv:2303.06615  (2023)

\bibitem{xu2019disn}
Xu, Q., Wang, W., Ceylan, D., Mech, R., Neumann, U.: Disn: Deep implicit surface network for high-quality single-view 3d reconstruction. Advances in neural information processing systems  \textbf{32} (2019)

\bibitem{yang2018unsupervised}
Yang, Z., Wang, P., Xu, W., Zhao, L., Nevatia, R.: Unsupervised learning of geometry from videos with edge-aware depth-normal consistency. In: Proceedings of the AAAI Conference on Artificial Intelligence. vol.~32 (2018)

\bibitem{yariv2021volume}
Yariv, L., Gu, J., Kasten, Y., Lipman, Y.: Volume rendering of neural implicit surfaces. Advances in Neural Information Processing Systems  \textbf{34},  4805--4815 (2021)

\bibitem{yariv2020multiview}
Yariv, L., Kasten, Y., Moran, D., Galun, M., Atzmon, M., Ronen, B., Lipman, Y.: Multiview neural surface reconstruction by disentangling geometry and appearance. Advances in Neural Information Processing Systems  \textbf{33},  2492--2502 (2020)

\bibitem{yi2023egolocate}
Yi, X., Zhou, Y., Habermann, M., Golyanik, V., Pan, S., Theobalt, C., Xu, F.: Egolocate: Real-time motion capture, localization, and mapping with sparse body-mounted sensors. arXiv preprint arXiv:2305.01599  (2023)

\bibitem{yu2022monosdf}
Yu, Z., Peng, S., Niemeyer, M., Sattler, T., Geiger, A.: Monosdf: Exploring monocular geometric cues for neural implicit surface reconstruction. arXiv preprint arXiv:2206.00665  (2022)

\bibitem{zhang2020visibility}
Zhang, J., Yao, Y., Li, S., Luo, Z., Fang, T.: Visibility-aware multi-view stereo network. arXiv preprint arXiv:2008.07928  (2020)

\bibitem{zhang2021learning}
Zhang, J., Yao, Y., Quan, L.: Learning signed distance field for multi-view surface reconstruction. In: Proceedings of the IEEE/CVF International Conference on Computer Vision. pp. 6525--6534 (2021)

\bibitem{zhang2021physg}
Zhang, K., Luan, F., Wang, Q., Bala, K., Snavely, N.: Physg: Inverse rendering with spherical gaussians for physics-based material editing and relighting. In: Proceedings of the IEEE/CVF Conference on Computer Vision and Pattern Recognition. pp. 5453--5462 (2021)

\end{thebibliography}
\end{document}

% TODO REVIEW: Replace with your title
\title{The Supplementary of ``Improving Neural Surface Reconstruction with Feature Priors from Multi-View Images''}

\titlerunning{Improving Neural Surface Reconstruction with Feature Priors}

% TODO FINAL: Replace with your author list. 
% Include the authors' OCRID for the camera-ready version, if at all possible.
\author{Xinlin Ren\inst{1}\orcidlink{0000-0002-8175-7392} \and
Chenjie Cao\inst{1,2}\orcidlink{0000-0003-3916-2843} \and
Yanwei Fu\inst{1}\orcidlink{0000-0002-6595-6893} \and
Xiangyang Xue\inst{1}\orcidlink{0000-0002-4897-9209}}

% TODO FINAL: Replace with an abbreviated list of authors.
\authorrunning{X. Ren et al.}
% First names are abbreviated in the running head.
% If there are more than two authors, 'et al.' is used.

% TODO FINAL: Replace with your institution list.
\institute{Fudan University\\ 
 \and Alibaba Group, DAMO Academy \\
 \email{ \{xlren20,cjcao20,yanweifu,xyxue\}@fudan.edu.cn }}

\maketitle

\section{Details of Patch-wise Consistent Loss}
\noindent\textbf{Normalization Cross Correlation (NCC).}
We can use NCC to measure the difference between patches of source view and reference view at the feature level:
\begin{equation}
\label{eq:ncc}
NCC(\textbf{F}_r(\hat{s}),\textbf{F}_s(\hat{s}'))=\mathrm{mean}\left(\frac{\mathrm{Cov}(\textbf{F}_r(\hat{s}), \textbf{F}_s(\hat{s}'))}{\sqrt{\mathrm{Var}(\textbf{F}_r(\hat{s}) \mathrm{Var}(\textbf{F}_s(\hat{s}')))}}\right),
\end{equation}
Where $\mathrm{Cov}$ and $\mathrm{Var}$ represent feature covariance and variance, respectively. $\hat{s}$ and $\hat{s}'$ denote the centers of reference and projected source patches\footnote{For simplicity, we use $\hat{s}$ and $\hat{s}'$ to present reference and projected source patches centered at $\hat{s}$, $\hat{s}'$, \emph{i.e.}, these patches are warped by the zero-cross distance of the related center points.}, respectively. $\textbf{F}_r$ and $\textbf{F}_s$ are the features of reference and source view.
Finally, we average all channel-wise feature correlations to the feature-level NCC.

\noindent\textbf{Structural Similarity (SSIM)} evaluates the local structural information in the images by comparing small patches of pixels. SSIM can also be applied to assess the difference between patches of the source view and reference view at the feature level:
% \begin{equation}
% \label{eq:ssim}
% \begin{split}
%     SSIM(\textbf{F}_r(\hat{s}),\textbf{F}_s(\hat{s}'))= \\
% \mathrm{mean}\left(\frac{(2\mu_{\textbf{F}_r}\mu_{\textbf{F}_s}+ c_1)(2\sigma_{\textbf{F}_r\textbf{F}_s}+c_2)  }{ (\mu_{\textbf{F}_r}^2+\mu_{\textbf{F}_s }^2+c_1)(\sigma_{\textbf{F}_r}^2+\sigma_{\textbf{F}_s }^2+c_2)}\right),
% \end{split}
% \end{equation}
% \begin{multline}
% SSIM(\textbf{F}_r(\hat{s}),\textbf{F}_s(\hat{s}'))= \\
% \mathrm{mean}\left(\frac{(2\mu_{\textbf{F}_r}\mu_{\textbf{F}_s}+ c_1)(2\sigma_{\textbf{F}_r\textbf{F}_s}+c_2)  }{ (\mu_{\textbf{F}_r}^2+\mu_{\textbf{F}_s }^2+c_1)(\sigma_{\textbf{F}_r}^2+\sigma_{\textbf{F}_s }^2+c_2)}\right),
% \end{multline}
\begin{equation}
\label{eq:ssim}
% \small
SSIM(\textbf{F}_r(\hat{s}),\textbf{F}_s(\hat{s}'))=\mathrm{mean}\left(\frac{(2\mu_{\textbf{F}_r}\mu_{\textbf{F}_s}+ c_1)(2\sigma_{\textbf{F}_r\textbf{F}_s}+c_2)  }{ (\mu_{\textbf{F}_r}^2+\mu_{\textbf{F}_s }^2+c_1)(\sigma_{\textbf{F}_r}^2+\sigma_{\textbf{F}_s }^2+c_2)}\right),
\end{equation}
where $\mu_{\textbf{F}_r}$, $\mu_{\textbf{F}_s}$ denote the mean of $\textbf{F}_r(\hat{s})$ and $\textbf{F}_s(\hat{s}')$. $\sigma_{\textbf{F}_r}$, $\sigma_{\textbf{F}_s}$ are the variance of these features, respectively. $\sigma_{\textbf{F}_r\textbf{F}_s}$ represents the covariance of  $\textbf{F}_r(\hat{s})$ and $\textbf{F}_s(\hat{s}')$. The constants $c_1$ and $c_2$ are set to 0.01 and 0.03, respectively, as suggested in~\cite{guo2022neural}. 
We average all channel-wise feature similarities as the SSIM.

\noindent\textbf{Patch Similarity (PS).} Similar to the pixel similarity, we use the average of  pixel similarity between corresponding elements in the two patches to represent the similarity of two patches as:
\begin{equation}
\label{eq:patch_similarity}
PS(\textbf{F}_r(\hat{s}),\textbf{F}_s(\hat{s}')) = \mathrm{mean}\left(\frac{\textbf{F}_r(\hat{s})\textbf{F}_s(\hat{s}')}{|\textbf{F}_r(\hat{s})|\  |\textbf{F}_s(\hat{s}')|}\right).
\end{equation}

\noindent\textbf{Patch-wise Consistent Loss.}
For the patch-wise consistent loss, we dynamically optimize each patch on 4 different source views with the lowest losses which are selected from 10 candidates.
% We select 10 source views and further choose 4 views of them with the best scores for computing the patch-wise consistent loss. 
This strategy is a simple yet effective way to mitigate occlusion as:
\begin{equation}
\label{eq:patch_consistent_loss}
\mathcal{L}_{patch} = \frac{\sum_{x=1}^N\sum_{s=1}^4 1-D_{patch}(\textbf{F}_r(x),\textbf{F}_s(x'))  }{4N},
\end{equation}
where $x$ and $x'$ denote the pixels and their corresponding pixels in the reference and source patches centered at $\hat{s}$ and $\hat{s}'$ respectively; and $N$ is the number of pixels in these patches. Besides, $D_{patch}$ can be represented as NCC, SSIM, and PS. Additionally, we use $11\times11$ patch size to compute the consistent loss.

\begin{table}[h]
\small
\vspace{-0.1in}
\centering
\caption{Quantitative results on the subset of DTU dataset. Features from pre-trained models are based on the \textbf{lowest} resolution.}
\label{tab:dtu_part_small}
% \small
\begin{tabular}{lccccc}
\toprule
Model        & Resolution  & case24 & case37 & case106 & Mean \\ 
\midrule
MAE ~\cite{he2022masked}        & $1024\times32\times48$  & 1.21   & 1.32   & 1.37    & 1.30  \\
ConvMAE~\cite{gao2022convmae}     & $768\times32\times48$   & 0.98   & 2.05   & 0.86    & 1.29 \\
CroCo~\cite{weinzaepfel2022croco} & $768\times32\times48$   & 1.19   & 2.05   & 0.82    & 1.35 \\
Twins~\cite{chu2021twins}      & $1024\times18\times24$ & 1.00      & 2.40    & 0.87    & 1.42 \\
ConvNeXt~\cite{liu2022convnet}  & $2048\times18\times24$  & 1.22   & 2.30    & 1.48    & 1.66 \\
SegFormer~\cite{xie2021segformer}  & $512\times18\times24$   & 0.99   & 2.44   & 0.97    & 1.46 \\
GLPN~\cite{kim2022global}  & $512\times18\times24$   & 0.97   & 1.31   & 0.88    & 1.05       \\
MiDaS~\cite{ranftl2020towards}        & $2048\times16\times24$  & 1.21   & 2.05   & 1.42    & 1.56 \\
IGEV~\cite{xu2023iterative}      & $160\times18\times24$   & 1.19   & 2.35   & 1.18    & 1.57 \\
QuadTree~\cite{tang2022quadtree}  & $256\times72\times96$   & 0.74   & 1.39   & 0.67    & 0.93 \\
CascadeMVS~\cite{gu2020cascade}    & $32\times256\times384$  &  0.79      &   1.24     &   1.00     &    1.01       \\
MVSFormer~\cite{cao2022mvsformer} & $32\times256\times384$  &  \textbf{0.65}      &   \textbf{1.00}     &    \textbf{0.63}    &  \textbf{0.76}    \\   
Baseline    &           /                  &  1.37  &  1.21  &  0.66   &  1.08 \\
\bottomrule
\end{tabular}
\vspace{-0.3in}
\end{table}

\begin{table}[h]
\small
\caption{Quantitative results on the subset of DTU dataset. Features from pre-trained models are based on the \textbf{highest} resolution.}
\label{tab:dtu_part_large}
\centering
% \small
\begin{tabular}{lccccc}
\toprule
Model        & Resolution  & case24 & case37 & case106 & Mean \\ 
\midrule

ConvMAE~\cite{gao2022convmae}      & $256\times128\times192$ & 0.99   & 1.68   & 0.81    & 1.16 \\
Twins~\cite{chu2021twins}      & $128\times128\times192$ & 0.89   & 1.52   & 0.87    & 1.09 \\
ConvNeXt~\cite{liu2022convnet}  & $256\times128\times192$ & 0.73   & 1.45   & 0.79    & 0.99 \\
SegFormer~\cite{xie2021segformer}    & $64\times128\times192$  & 0.83   & 1.25   & 0.88    & 0.98 \\
GLPN~\cite{kim2022global}     & $64\times128\times192$  & 0.90    & 1.32   & 0.81    & 1.01 \\
IGEV~\cite{xu2023iterative}         & $96\times128\times192$  & 0.88   & 1.23   & 0.76    & 0.95 \\
RAFT-Stereo~\cite{lipson2021raft} & $256\times128\times192$ & 0.59   & 0.92   & 0.66    & 0.72 \\
QuadTree~\cite{tang2022quadtree}     & $128\times256\times384 $ & 0.48   & 0.85   & 0.55    & 0.62 \\
CascadeMVS~\cite{gu2020cascade}   & $8\times512\times768$   & 0.46       &    0.75      &  0.55       &  0.59    \\
MVSFormer~\cite{cao2022mvsformer}    & $8\times512\times768$   &  \textbf{0.40}      &  \textbf{0.65}    &   \textbf{0.49}      &  \textbf{0.51}     \\
Baseline     &            /                 &  1.37  &  1.21  &  0.66   &  1.08 \\
\bottomrule
\end{tabular}
\end{table}

\section{Implementation Details}
For a fair comparison, we adopt the same architecture as the current works~\cite{fu2022geo,wang2021neus,chen2023recovering}. The geometry network consists of an 8-layer MLP with 256 hidden units, which is initialized using the geometric initialization proposed in~\cite{atzmon2020sal}. And it outputs SDF and a 256-dimension feature vector which will be concatenated with normal, view direction and 3D point to regress color. 
The radiance net has a 4-layer MLP with 256 hidden units, and positional encoding with 6 and 4 frequencies is applied to encode the 3D position and view direction, respectively. Additionally, we sample 512 rays  per batch which follow the hierarchical sampling strategy in NeuS~\cite{wang2021neus} where each ray has 64 coarse and 64 fine sampled points. Besides, the background image is modeled using 32 points outside the unit sphere, following the approach of Nerf++~\cite{zhang2020nerf++}. We train the network for 300k iterations costing 10 hours on a single A6000.

\section{Detailed Quantitative Results for Different Pretext Tasks }
We evaluate models of various pretext tasks by applying pixel similarity as the consistency loss on a subset of the DTU dataset including scan24, scan37, and scan106.
These objects are relatively challenging due to difficult lighting conditions and occlusions. Tab.~\ref{tab:dtu_part_small} and Tab.~\ref{tab:dtu_part_large} show the lowest and highest features of various models, respectively.

\section{Additional Results of Feature Priors on 
Grid-Based Representation}
\begin{figure}
\begin{center}
\includegraphics[width=0.9\linewidth]{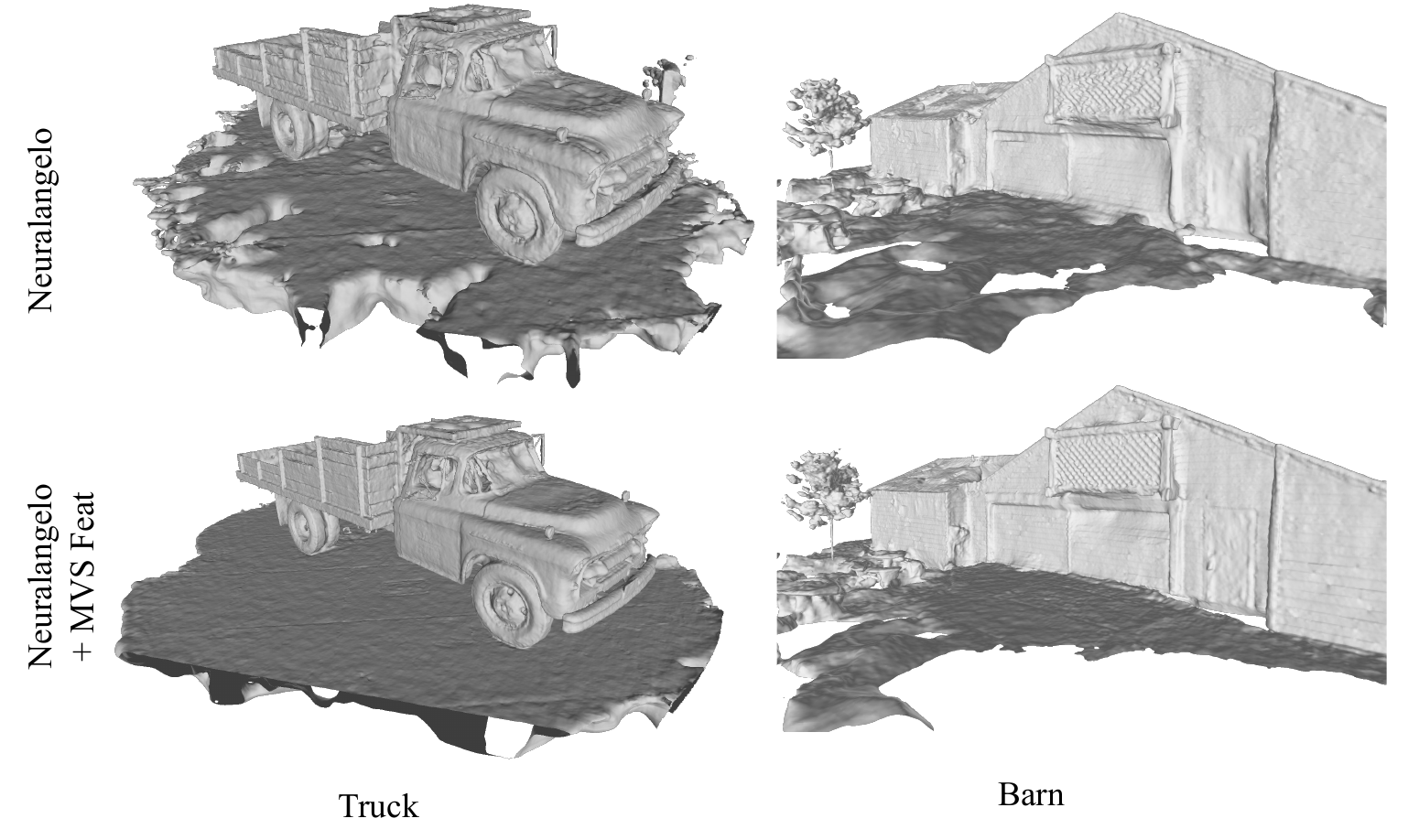}
\vspace{-0.1in}
 \caption{Qualitative results of Neuralangelo with feature priors on Tanks\&Temples dataset.}
\label{fig:supp_outdoor}
\end{center}
\vspace{-0.3in}
\end{figure}

We incorporate feature priors into grid-based Neuralangelo~\cite{li2023neuralangelo} on DTU~\cite{jensen2014large} and Tanks\&Temples\cite{knapitsch2017tanks}
as in Tab.~\ref{tab:supp_dtu_prior} and Fig.~\ref{fig:supp_outdoor}.

\begin{table}
\small
\vspace{-0.2in}
\caption{Quantitative results of Neuralangelo with feature priors on DTU dataset.}
\label{tab:supp_dtu_prior}
\resizebox{0.95\linewidth}{!}{
\begin{tabular}{ccccccccccccccccc}
\toprule 
  Scan    & 24 & 37 & 40 & 55 & 63 & 65 & 69 & 83 & 97 & 105 & 106 & 110 & 114 & 118 & 122 & Mean\tabularnewline
\hline 
Neuralangelo & 0.50 & 1.03 & 0.83 & 0.47 & 2.93 & 0.61 & 2.49 & 2.72 & 3.31 & 2.87 & 1.69 & 2.98 & 0.77 & 4.16 & 2.80 & 2.01\tabularnewline
+ MVS Feat & 0.52 & \textbf{0.82} & \textbf{0.46} & \textbf{0.38} & \textbf{1.60} & 0.63 & \textbf{1.15} & \textbf{1.56} & \textbf{1.89} & \textbf{0.85} & \textbf{0.70} & \textbf{0.61} & \textbf{0.30} & \textbf{1.23} & \textbf{0.85} & \textbf{0.90}\tabularnewline
\bottomrule
\end{tabular}}
\vspace{-0.2in}
\end{table}

\subsection{Quantitative results of Feature Priors on DTU dataset}
We train Neuralangelo with MVSFormer feature priors on the DTU dataset for 300k iterations. As indicated in Tab.~\ref{tab:supp_dtu_prior}, the integration of MVSFormer features results in a reduction in errors.

\subsection{Qualitative Results of Feature Priors on Outdoor Scenes}
We train Neuralangelo with MVSFormer feature priors on the Tanks\&Temples dataset for 300k iterations. From Fig.~\ref{fig:supp_outdoor}, Neuralangelo with feature priors results in a smoother surface and more accurate geometry.

\subsection{Efficiency of Feature Priors}
\begin{figure}
\begin{center}
\includegraphics[width=1.0\linewidth]{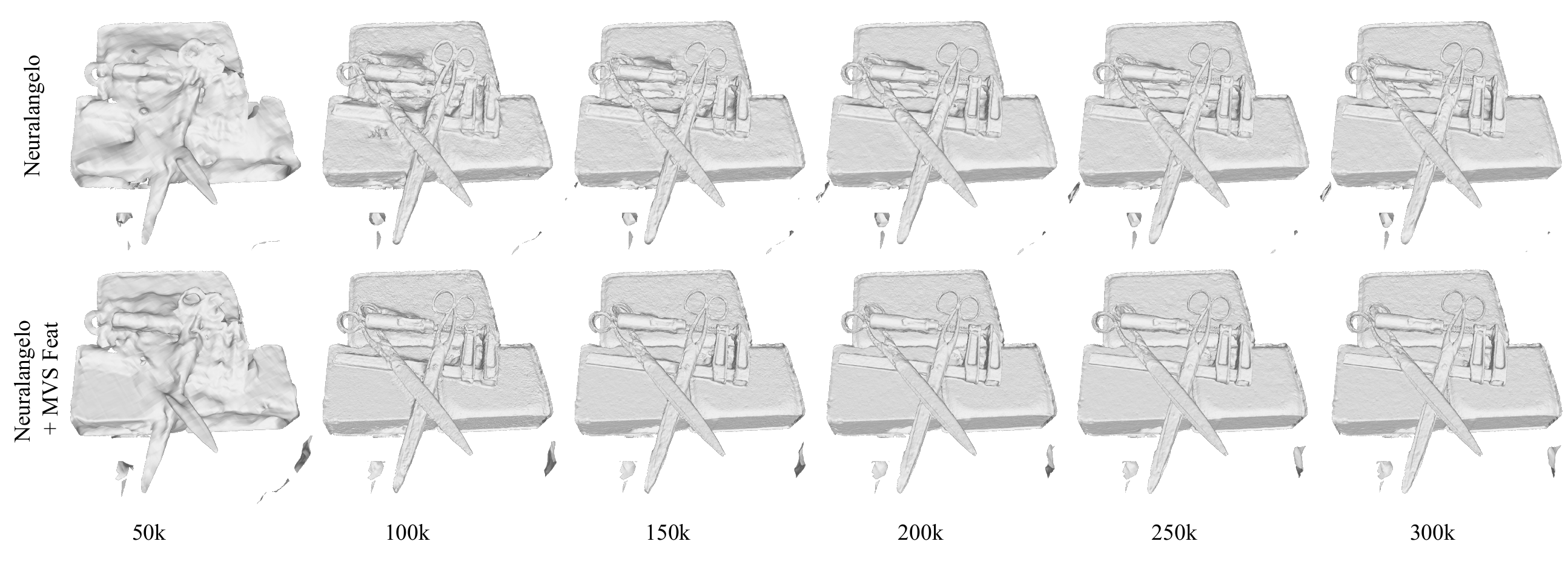}
\vspace{-0.1in}
 \caption{Qualitative results with feature priors  Scan37 at different training iterations. }
\label{fig:supp_efficiency}
\end{center}
\vspace{-0.3in}
\end{figure}

We present the outcomes of Neuralangelo incorporating feature priors on Scan37 across various training iterations in Fig.~\ref{fig:supp_efficiency}.
Remarkably, the results achieved by Neuralangelo with feature priors at 150k iterations surpass the accuracy of Neuralangelo without feature priors even after 300k iterations. This underscores the significant impact of feature priors in enhancing network convergence.

\begin{figure}
\begin{center}
\includegraphics[width=0.9\linewidth]{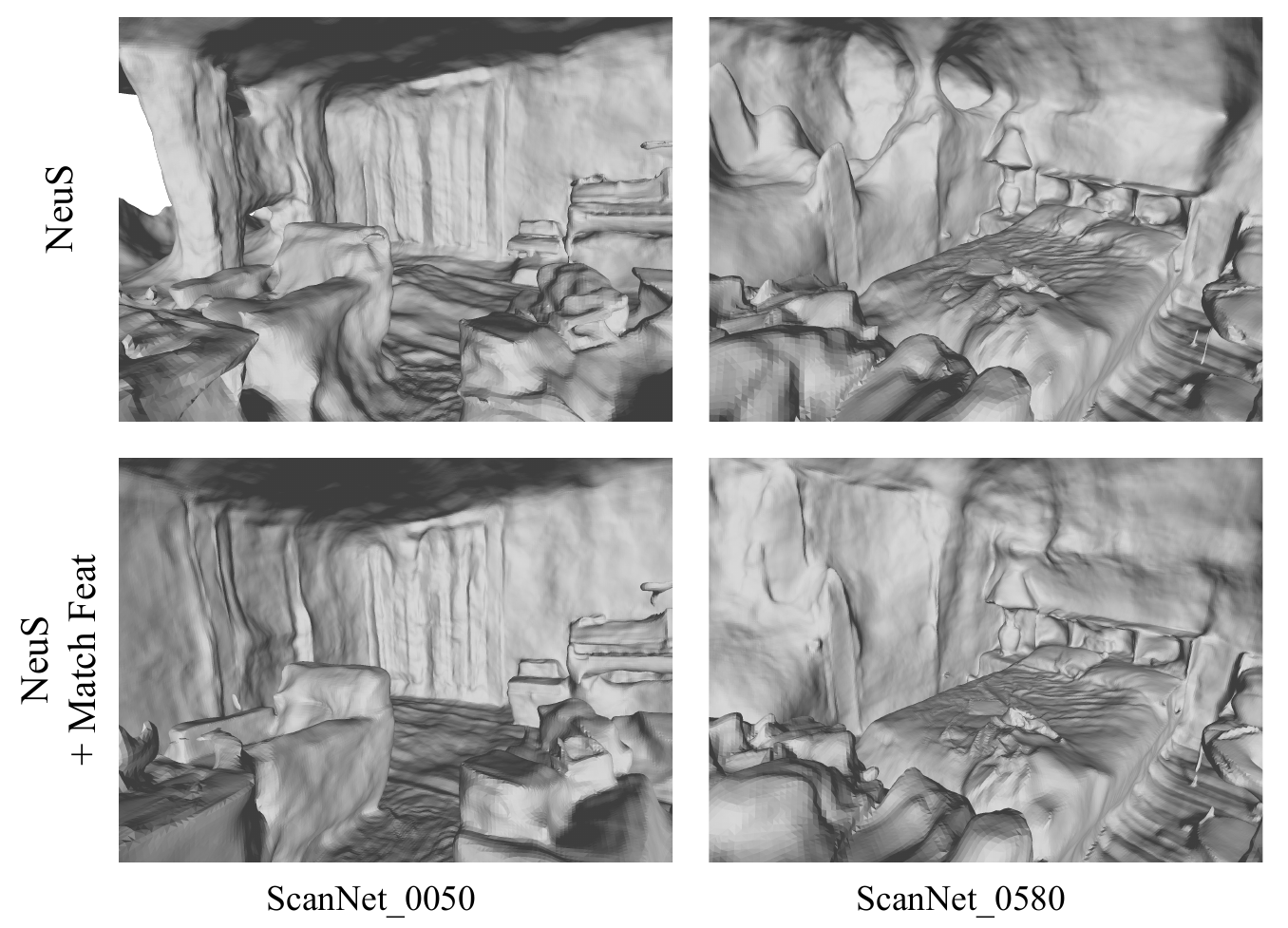}
\vspace{-0.1in}
 \caption{Qualitative results of NeuS with feature priors on ScanNet dataset. }
\label{fig:supp_indoor}
\end{center}
\vspace{-0.3in}
\end{figure}

\section{Results of Feature Priors on 
Indoor Scenes}
We provide more qualitative results on ScanNet of NeuS~\cite{wang2021neus} with QuadTree~\cite{tang2022quadtree} Feature in Fig.~\ref{fig:supp_indoor}. Notably, NeuS with match features can reconstruct more accurate geometry than NeuS, verifying that our method can be also generalized to indoor scenes.

\section{More Analysis}

\subsection{GPU Memory Consumption}
\label{sec:supp_gpu}
We show the GPU memory cost of our Match-NeuS and MVS-NeuS in Tab.~\ref{tab:gpu_memory}. 
To improve efficiency, we save all off-the-shelf features in the GPU at once, avoiding costly GPU and CPU memory swappings.
Compared with the baseline, this strategy costs extra GPU memory consumption.
But we should clarify that these memory costs are optional to accelerate the training phase without frequently loading features into the GPU devices.

\begin{table}[]
\vspace{-0.2in}
\caption{GPU memory consumption on DTU dataset.
$\dag$ denotes saving all features in the GPU beforehand.}
    \centering
    \resizebox{0.98\linewidth}{!}{
    \begin{tabular}{ccccc}
    \toprule
        Model & Pre-Trained & Consistent Loss & GPU Memory (MB) &  GPU Memory$^\dag$ (MB)\\ \midrule
        Baseline &   /  &   /  &  7033 & 7033 \\\midrule
        \multicolumn{1}{c}{\multirow{4}{*}{Match-NeuS}} & \multicolumn{1}{c}{\multirow{4}{*}{QuadTree~\cite{tang2022quadtree}}}   &   Pixel Similarity & 8415 &18999  \\ 
        \multicolumn{1}{c}{}  &   &Patch Similarity & 13823 &24401  \\
        \multicolumn{1}{c}{}  &   &Patch NCC  & 14097 &24683 \\
        \multicolumn{1}{c}{}  &   &Patch SSIM & 14373 &24955 \\ \midrule
        \multicolumn{1}{c}{\multirow{4}{*}{MVS-NeuS}} & \multicolumn{1}{c}{\multirow{4}{*}{MVSFormer~\cite{cao2022mvsformer}}}   &   Pixel Similarity & 7443 & 10089  \\ 
        \multicolumn{1}{c}{}  &   &Patch Similarity & 8126 &10724 \\
        \multicolumn{1}{c}{}  &   &Patch NCC & 8303 & 10951 \\
        \multicolumn{1}{c}{}  &   &Patch SSIM & 8327 & 10949 \\ \bottomrule
    \end{tabular}
    
    \label{tab:gpu_memory}}
    
\end{table}

\subsection{The Effect of Patch Size }
In our paper, we use $11\times11$ patch size as default to compute the consistent loss. Here we show the performance of NCC with different patch sizes on Scan37 of the DTU dataset. The result in Tab.~\ref{tab:patch_size} shows that using NCC with patch sizes that are too small results in poor performance. 
Conversely, using too large patch sizes may lead to slightly inferior results. The best performance is achieved with $7\times7$ and $11\times11$ patches. It is reasonable that a small patch size suffers from the local minimum while too large patch sizes can't capture detailed geometry accurately.
\begin{table}[]
\small
\caption{ Quantitative results of NCC with different patch sizes on scan37 of DTU Dataset.}
    \centering
    \vspace{-0.2in}
    \begin{tabular}{cccc}
    \toprule
        Model & Pre-Trained & Patch Size & Mean\\ \midrule
        Baseline &   /  &   /  &  1.21  \\\midrule
       \multicolumn{1}{c}{\multirow{4}{*}{Match-NeuS}} & \multicolumn{1}{c}{\multirow{4}{*}{QuadTree~\cite{tang2022quadtree}}}   &  $3\times3$ & 0.906\\
        \multicolumn{1}{c}{}  &   & $7\times7$ & 0.684   \\
        \multicolumn{1}{c}{}  &   & $11\times11$  &  0.671 \\
        \multicolumn{1}{c}{}  &   & $15\times15$ & 0.706  \\ 
        \multicolumn{1}{c}{}  &   & $19\times19$ & 0.713 \\ \midrule
        \multicolumn{1}{c}{\multirow{4}{*}{Match-NeuS}} & \multicolumn{1}{c}{\multirow{4}{*}{MVSFormer~\cite{cao2022mvsformer}}}   &  $3\times3$ & 0.828 \\
        \multicolumn{1}{c}{}  &   & $7\times7$ & \textbf{0.602}   \\
        \multicolumn{1}{c}{}  &   & $11\times11$  & 0.631  \\
        \multicolumn{1}{c}{}  &   & $15\times15$ & 0.653  \\ 
        \multicolumn{1}{c}{}  &   & $19\times19$ & 0.678 \\ \bottomrule
        
    \end{tabular}
    \vspace{-0.2in}
    \label{tab:patch_size}
    
\end{table}
\subsection{NCC with Multi-Scale Features }
We show additional quantitative results of NCC with multi-scale features on scan37 of the DTU dataset in  Tab.~\ref{tab:multi_scale}. Using multi-scale features leads to worse results than only using the highest-resolution features.

\subsection{Combination of MVS and Image Matching features}
We show additional quantitative results of combining MVS and image matching features as mixed features in Tab.~\ref{tab:supp_dtu_combine}. The method employing patch NCC with these mixed features is denoted as MVS-Match-NeuS. From Tab.~\ref{tab:supp_dtu_combine}, we observe that the integration of MVS and image matching features does not consistently enhance performance. Moreover, it incurs a higher memory cost to store all these features in the GPU as illustrated in Sec.~\ref{sec:supp_gpu}.

\begin{table}
\small
\centering
\caption{Results of CD (mm) compared on DTU dataset (lower is better). The best results are in \textbf{bold}.\label{tab:supp_dtu_combine}}
\vspace{-0.1in}
 \resizebox{1.0\linewidth}{!}{
\begin{tabular}{ccccccccccccccccc}
\toprule
Scan & 24 & 37 & 40 & 55 & 63 & 65 & 69 & 83 & 97 & 105 & 106 & 110 & 114 & 118 & 122 & Mean\tabularnewline \midrule

MVS-NeuS & \textbf{0.37} & \textbf{0.63} & \textbf{0.31} & \textbf{0.32} & \textbf{0.79} & \textbf{0.51} & 0.50 & \textbf{1.12} & \textbf{0.87} & 0.62 & 0.46 & \textbf{0.64} & \textbf{0.29} & \textbf{0.38} & \textbf{0.41} & \textbf{0.548}\tabularnewline
% \hline 
Match-NeuS & 0.39 & 0.67 & 0.32 & 0.33 & \textbf{0.79} & \textbf{0.51} & 0.49 & 1.21 & \textbf{0.87} & 0.63 & 0.48 & 0.71 & \textbf{0.29} & \textbf{0.38} & \textbf{0.41} & 0.565\tabularnewline
% \hline 
MVS-Match-NeuS & \textbf{0.37} & \textbf{0.63} & 0.33 & \textbf{0.32} & 0.80 & 0.52 & \textbf{0.46} & 1.16 & 0.90 & \textbf{0.60} & \textbf{0.45} & 0.74 & \textbf{0.29} & 0.40 & 0.42 & 0.560\tabularnewline
\bottomrule
\end{tabular}
\par}
\end{table}

% We show additional quantitative results of NCC with multi-scale features on scan37 of the DTU Dataset in  Tab.~\ref{tab:multi_scale}. Using multi-scale features leads to worse results than only using the highest-resolution features.

\begin{table}[]
\caption{Quantitative results of NCC with multi-scale features on scan37 of DTU Dataset.}
% Models without multi-scale features are using the highest resolution features and the baseline doesn't apply any consistent loss. }
    \centering
    \small
    \begin{tabular}{cccc}
    \toprule
        Model & Pre-Trained  & Multi-Scale Features &Mean\\ \midrule
        Baseline &   /  &   /  &  1.21  \\\midrule
        \multicolumn{1}{c}{\multirow{2}{*}{Match-NeuS}} & \multicolumn{1}{c}{\multirow{2}{*}{QuadTree~\cite{tang2022quadtree}}}& $\times$ &  0.671 \\
         \multicolumn{1}{c}{}  &   &  $\checkmark$ & 0.753   \\ \midrule
         \multicolumn{1}{c}{\multirow{2}{*}{MVS-NeuS}} & \multicolumn{1}{c}{\multirow{2}{*}{MVSFormer~\cite{cao2022mvsformer}}}& $\times$ &  \textbf{0.631} \\
        \multicolumn{1}{c}{}  &   &  $\checkmark$ & 0.676   \\ \midrule      
    \end{tabular} 
    \label{tab:multi_scale}
 
\end{table}

\section{Additional Qualitative Results}
\begin{figure}
\begin{center}
\includegraphics[width=1.0\linewidth]{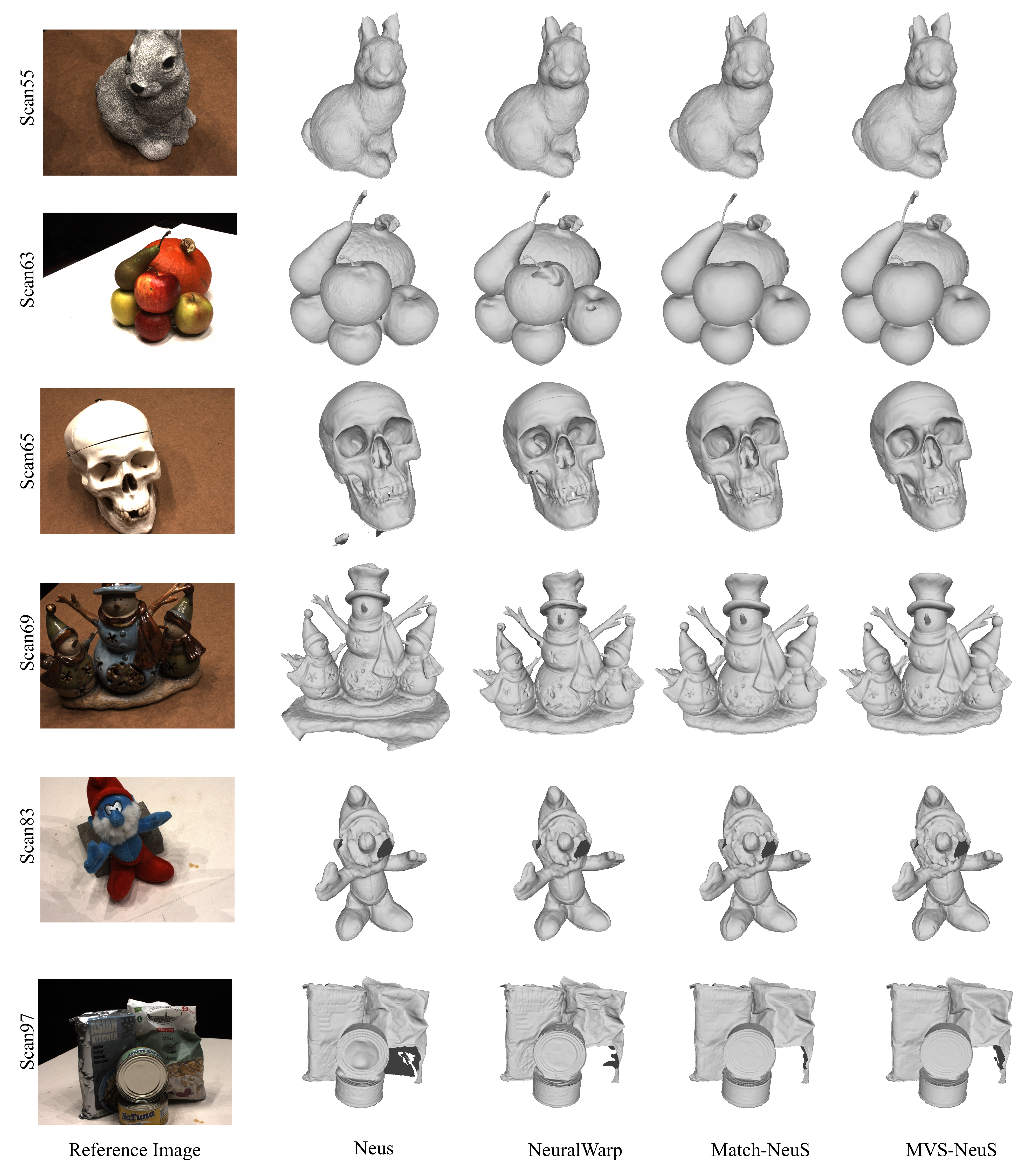}
\vspace{-0.1in}
 \caption{More qualitative results on the DTU dataset.}
\label{fig:supp_dtu1}
\end{center}
\vspace{-0.15in}
\end{figure}

\begin{figure}
\begin{center}
\includegraphics[width=1.0\linewidth]{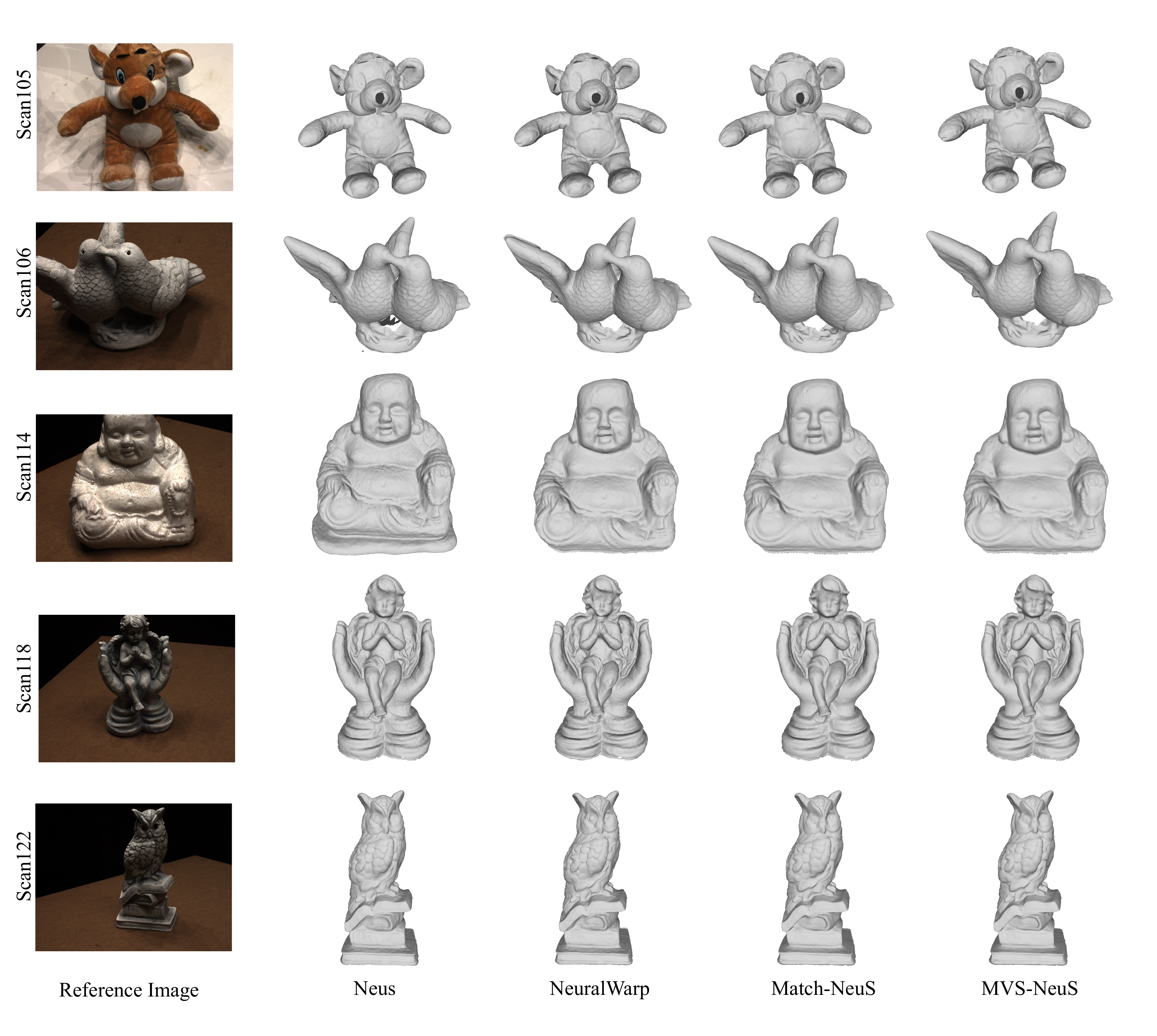}
\vspace{-0.1in}
 \caption{More qualitative results on the DTU dataset.}
\label{fig:supp_dtu2}
\end{center}
\vspace{-0.15in}
\end{figure}

We show additional qualitative results on the DTU dataset in Fig.~\ref{fig:supp_dtu1}, Fig.~\ref{fig:supp_dtu2}.

\section{Limitation}
As shown in Tab.~\ref{tab:gpu_memory}, we need to save all off-the-shelf features in the GPU at once, avoiding costly GPU and CPU memory swappings. This becomes particularly crucial when utilizing NCC with high-resolution features, such as QuadTree, which have large dimensions and require more GPU memory. Another limitation is that our work is based on NeuS~\cite{wang2021neus} and it is difficult to model transparent surfaces. 

\section{Board Impact}
Our work focuses on leveraging the wealth of 2D image data to learn and reconstruct the 3D geometry of the world. Our approach enables high-quality 3D reconstruction of objects  using only standard images. The applications of our algorithm are diverse and can be utilized in various fields where 3D information is essential but only 2D images are available. This has the potential to benefit areas such as manufacturing, virtual reality, surveillance, healthcare, and many others.

% {
%     \small
%     \bibliographystyle{ieeenat_fullname}
%     \bibliography{main}
% }

% WARNING: do not forget to delete the supplementary pages from your submission 
% \input{sec/X_suppl}
\bibliographystyle{splncs04}
\bibliography{main}